\documentclass[preprint,12pt]{elsarticle}

\usepackage{morefloats}
\usepackage{flushend}
\usepackage{amsfonts}
\usepackage[cmex10]{amsmath}
\usepackage{algorithm}
\usepackage{algorithmic}
\usepackage{multirow}

\usepackage[tight,footnotesize]{subfigure}
\usepackage{tikz}
\usetikzlibrary{arrows,shapes,shadows,positioning,fadings,decorations,decorations.pathmorphing}
\tikzstyle{block}=[draw opacity=0.7,line width=1.4cm]

\newcommand{\node}{v}

\newcommand{\labelfun}{L}

\usepackage{ulem}
\usepackage{color}
\newcommand{\vecchio}[1]{\textcolor{blue}{}}
\newcommand{\nuovo}[1]{#1}
\newcommand{\rosso}[1]{#1}
\newcommand{\nero}[1]{\textcolor{black}{#1}}

\definecolor{darkblue}{rgb}{0,0,0.75}
\journal{Neurocomputing}

\begin{document}
\begin{frontmatter}



\title{An Empirical Study on Budget-Aware Online Kernel Algorithms for Streams of Graphs}
\author[qcri]{Giovanni Da San Martino}
  \ead{gmartino@qf.org.qa}
\author[unipd]{Nicol\`o Navarin\corref{cor1}}
  \ead{nnavarin@math.unipd.it}
\author[unipd]{Alessandro Sperduti} 
  \ead{sperduti@math.unipd.it}

\cortext[cor1]{Corresponding author}
\address[qcri]{Qatar Computing Research Institute, HBKU, P.O. Box 5825 Doha, Qatar}
\address[unipd]{Department of Mahematics, University of Padova, via trieste 63, Padova, Italy}
\begin{abstract}

Kernel methods are considered an effective technique for on-line learning. 
Many approaches have been developed for compactly representing the dual solution of a kernel method when the problem imposes memory constraints.  
However, in literature no work is specifically tailored to streams of graphs. 
Motivated by the fact that the size of the feature space representation of many state-of-the-art graph kernels is relatively small and thus it is explicitly computable, we study whether executing kernel algorithms in the feature space can be more effective than the classical dual approach.   
We study three different algorithms and various strategies for managing the budget. 
Efficiency and efficacy of the proposed approaches are experimentally assessed on relatively large graph streams exhibiting concept drift. 
It turns out that, when strict memory budget constraints have to be enforced, working in feature space, given the current state of the art on graph kernels, is more than a viable alternative to dual approaches, both in terms of speed and classification performance. 

\end{abstract}


\end{frontmatter}

\section{Introduction}

The amount of data generated in different areas by computer systems is growing at an extraordinary pace, mainly due to the advent of 
technologies related to the web, ubiquitous services and embedded systems that aim at monitoring the environment in which they are
immersed in. 
Data are, in some cases, generated at a constant rate by sources that can potentially emit an unbounded sequence of elements, i.e. data streams. 
The processing of data streams requires special care from a computational point of view, since only bounded time and memory resources might be available. Indeed, online algorithms may be required to scale linearly with the number of data items and use a constant, a priori determined, amount of memory (budget). 
An example of a learning task on streams is binary classification, where the goal is to approximate a function $f:\mathbb{X}\rightarrow \{-1,1\}$ which partitions the input domain $\mathbb{X}$ into two classes. 
 When dealing with streams, it was early recognized that they tend to evolve with time, giving rise to the well known {\it concept drift}  phenomenon \cite{klinkenberg2004}, which consists in the function $f()$ changing over time. 

In this paper, we focus on graph streams, which involve a large range of application tasks such as
 chemical compound or image classification (see Sections~\ref{sec:chemicaldataset} and \ref{sec:imagedataset}, respectively),
 as well as malware detection \cite{citeulike:10456962}, where executables codes represent graph nodes and control flow instructions and API calls represent edges, and  Fault Diagnosis in Sensor Networks \cite{alippi}.
Note that we assume that the source generating the stream emits one graph at a time (i.e.,  we do not have an edge stream as, for example, in \cite{Aggarwal11}). 

\nero{The traditional approach when dealing with structured data is to transform the data into a suitable vectorial representation. When the examples are graphs, the mapping is commonly referred to as  graph embedding \cite{GIBERT2013}. The drawbacks of this approach are that the embedding is task-dependent, and generally computationally expensive. Moreover, the dimensionality of the vector in which the mapping is performed has to be fixed a-priori (see e.g.~\cite{Bianucci2000}), and it is the same for all examples ignoring the differences in the intrinsic complexity of each graph. 
}

\nero{
A viable alternative to graph embedding is  the application of graph kernel methods, which is the approach we consider in this paper.}
Kernel methods are considered state of the art techniques for classification tasks \cite{cristianini2000introduction, hofmann2008,DaSanMartino2010,Dasan2012}. 
The class of kernel methods comprises all those learning algorithms that do not require an explicit representation of the inputs but only information about the similarity between them. 
The primal version of a kernel method maps the data onto a vectorial feature space \nero{(possibly infinite-dimensional)}: the similarity can be expressed as a dot product in such space. Any kernel method has a correspondent dual version in which each dot product in feature space is replaced by the evaluation of a correspondent kernel function defined on the input space. 
 The great advantage of kernel methods is the fact that the space and time complexity depends on the kernel function and not on the size of the corresponding feature space. 
Consequently, the size of the model, i.e. the space needed by the learning algorithm for representing its current solution, is defined in terms of a subset of input examples instead of a subset of features. 
 It is recognized that, when the model is expressed as a set of examples, its size tends to grow proportionally to the number of instances emitted by the stream \cite{Aiolli2006}. Various approaches have been defined to limit the size of the model \cite{Kivinen2004, Crammer_Dekel_Keshet_Shalev-Shwartz_Singer_2006, DBLP:journals/jmlr/OrabonaKC09}. However, their application to graph data has been practically limited due to the fact that kernels for graphs tend to be computationally very expensive \cite{Gartner2003a, Borgwardt2005, Mah'e2009}. 
Recently a few kernels for graphs have been defined which are both efficient and have very competitive performances on many benchmark datasets \cite{Shervashidze2009a, Costa2010, Dasan2012}. Their complexity ranges from linear in the number of  edges \cite{Shervashidze2009a} to a logarithmic factor above linear in the number of nodes \cite{Dasan2012}, thus they might be ideal candidates for being employed on data streams.  One of their key characteristics is that they lead to models that can be represented compactly in the primal space. Thus, for these kernels, both techniques defined for the primal and dual space can be effectively exploited. 

The main goal of the paper is to study which of the two approaches is best suited for graph streams. 
We empirically study the behavior of three different algorithms defined in the primal or in the dual space, using the state-of-the-art graph kernels described in \cite{Shervashidze2009a, Costa2010, Dasan2012} and with multiple techniques for managing the budget.
We show experimental results on reasonably large real-world datasets and in the presence of a (controlled) concept drift. The results suggest that, under specific budget constraints, working in the primal space is faster and leads to better or comparable results with respect to the classic dual approach.

\nero{The paper is organized as described in the following.
Section~\ref{sec:background}, after introducing some notation, recalls important background notions for understanding the paper: graph kernels, online learning algorithms on a budget defined in primal or dual space.
Section~\ref{sec:budgetalgorithms} extends the previously presented online learning algorithms to graph data and discusses several model-pruning strategies to ensure that strict budget constraints are satisfied.
Section~\ref{sec:experiments} studies the performances of the learning algorithms on a budget with respect to the various model strategies and kernel functions. Finally, Section~\ref{sec:conclusions} draws conclusions.}


\section{Background} \label{sec:background}

This section introduces the concepts and algorithms used in the remainder of the paper. 
\nero{We start by introducing some notation in Section~\ref{sec:notation}.}
Section~\ref{sec:oddkernel} briefly reviews kernel functions for graphs outlining the fact that some of the state-of-the-art ones have both low computational complexity and a compact representation as a set of features. Motivated by this last observation, we describe state-of-the-art kernel methods for online learning and budget management techniques working in the dual space, in Section~\ref{sec:dualkernels},  and  online learning algorithms working directly in feature space, in Section~\ref{sec:primalalgs}. 
\subsection{Notation} \label{sec:notation}
A graph $G(V,E,\labelfun)$ is a triplet where $V$ is the set of vertices, $E$ the set of edges and $\labelfun()$ a function mapping nodes to a set of labels $A$. 
A proper subgraph $G_2=(V_2,E_2,\labelfun)$ of $G_1=(V_1,E_1,\labelfun)$ is a graph for which $V_2\subseteq V_1$, $E_2=E_1\cap (V_2\times V_2)$.
A directed acyclic graph (DAG)  is a graph where edges are directed and no directed cycle is present. 
A {\it proper rooted substructure} of a DAG  $D$ is defined in this paper as a subgraph of $D$ obtained by considering a node $\node$ of $D$ and all the nodes which can be reached from $\node$ using the directed edges of $D$. 
A tree is a directed acyclic graph where each node has at most one incoming edge. 
A proper subtree rooted at node $\node$ comprises $\node$ and all its descendants. 
We denote with $\rho$ the maximum outdegree of a graph. 
\subsection{Graph Kernels}
\label{sec:oddkernel}

In order to apply a kernel method to graph data, an appropriate kernel function must be provided. 
Such function, defined on any pair of instances of a domain must be symmetric positive semidefinite. 
Various similarity measures can be exploited to define a kernel for graphs. For example, a similarity score can be given by the number of subgraphs that two graphs $G_1$ and $G_2$ share. 
Unfortunately, the implementation of this simple idea is very expensive from a computational point of view since recognizing if a subgraph $g_1$ of $G_1$ is isomorphic to a subgraph $g_2$ of $G_2$ requires to solve a {\it subgraph isomorphism problem}, which is known
to be NP-Complete \cite{Gartner2003a}. 

Most of the research on graph kernels proceeded by focusing on a restricted class of substructures for which the membership to a graph can be decided in polynomial time (e.g., walks \cite{Gartner2003a, Kashima2003, Vishwanathan2006}, shortest paths \cite{Borgwardt2005, Heinonen2012}, subtree patterns \cite{Mah'e2009}, small-sized subgraphs \cite{Shervashidze2009}) with the aim of obtaining a feature space as large as possible. However, the complexity of the cited algorithms spans from $O(n^3)$ to $O(n^{6})$\footnote{\nero{The kernel in \cite{Shervashidze2009} can be computed in $O(n\rho^{k-1})$, where $k$ is the size of the considered subgraphs, on unlabeled graphs. However, in this paper we deal with labeled graphs and the complexity of the kernel for this case is $O(n^k)$.}}, where $n$ is the size (number of nodes) of the graphs, which make them hardly applicable to on-line learning tasks with strict time constraints. 

Recently, a few kernels with complexity $O(m)$, where $m$ is the number of edges, or $O(n \log n)$, have been defined \cite{Shervashidze2009a, Costa2010, Dasan2012}. Despite their low complexity their performance is considered state of the art on many benchmark datasets. Moreover, their low complexity allow them to be applied to very large datasets. 
The \textit{Weisfeiler-Lehman subtree kernel} \cite{Shervashidze2009a} considers the number of subtree patterns (subtrees where every node in the original graph may appear multiple times) up to a fixed  height $h$. This kernel can be computed in $O(hm)$ time on a pair of graphs $G_1$ and $G_2$, where $m=\max(|E_1|, |E_2|)$. 
\rosso{Note that the $h$ is a kernel parameter and the authors always use a constant value, so the complexity practically is $O(m)$. }
The \textit{Neighborhood subgraph pairwise distance kernel} (NSPDK) \cite{Costa2010} decomposes a graph into pairs of small subgraphs \nero{of radius at most $h$,} up to a maximum distance $d$: every feature in the explicit feature space represents two particular subgraphs being at a certain distance. \rosso{Here $d$ and $h$ are kernel parameters which, in order to reduce the computational burden of the kernel evaluation, in practice are kept constant~\cite{Costa2010}. }
Finally, the \textit{ODD$_{ST}$} kernel, a member of the Ordered Decompositional DAGs Kernel family for graphs \cite{Dasan2012}, decomposes a graph of $n$ nodes into $n$ DAGs. Each DAG is obtained performing a breadth first visit of the graph, up to a fixed height $h$ set by the user, and removing the nodes inducing a cycle. The features associated with a graph are the proper rooted substructures of each DAG. 

The set of non-zero features related to the \textit{Weisfeiler-Lehman subtree}, the \textit{Neighborhood subgraph pairwise distance} and the \textit{ODD$_{ST}$} kernels, and consequently the associated models, tend to have a compact representation. 
The number of features generated for a graph is {\it at most}:  \mbox{$n h$} for the \textit{Weisfeiler-Lehman subtree kernel} \cite{Shervashidze2009a}; $\frac{hn\rho^d}{2}$ for NSPDK, where $\frac{n\rho^d}{2}$ is an upper bound on the number of pairs of nodes that are at most at distance $d$; $n\rho^h$ for \textit{ODD}$_{ST}$ \cite{Dasan2012}. 

\rosso{Note that the kernel parameters $h$, $d$ are assumed to be constant \cite{Shervashidze2009a, Costa2010, Dasan2012} and that, in many practical applications, $\rho$ can be considered constant as well, thus the number of features generated by the different kernels is practically linear.}
This property will be exploited by the online learning algorithms described in Section~\ref{sec:primalalgs}. 

\begin{figure}[t]
\centering
 \includegraphics[width=0.8\linewidth]{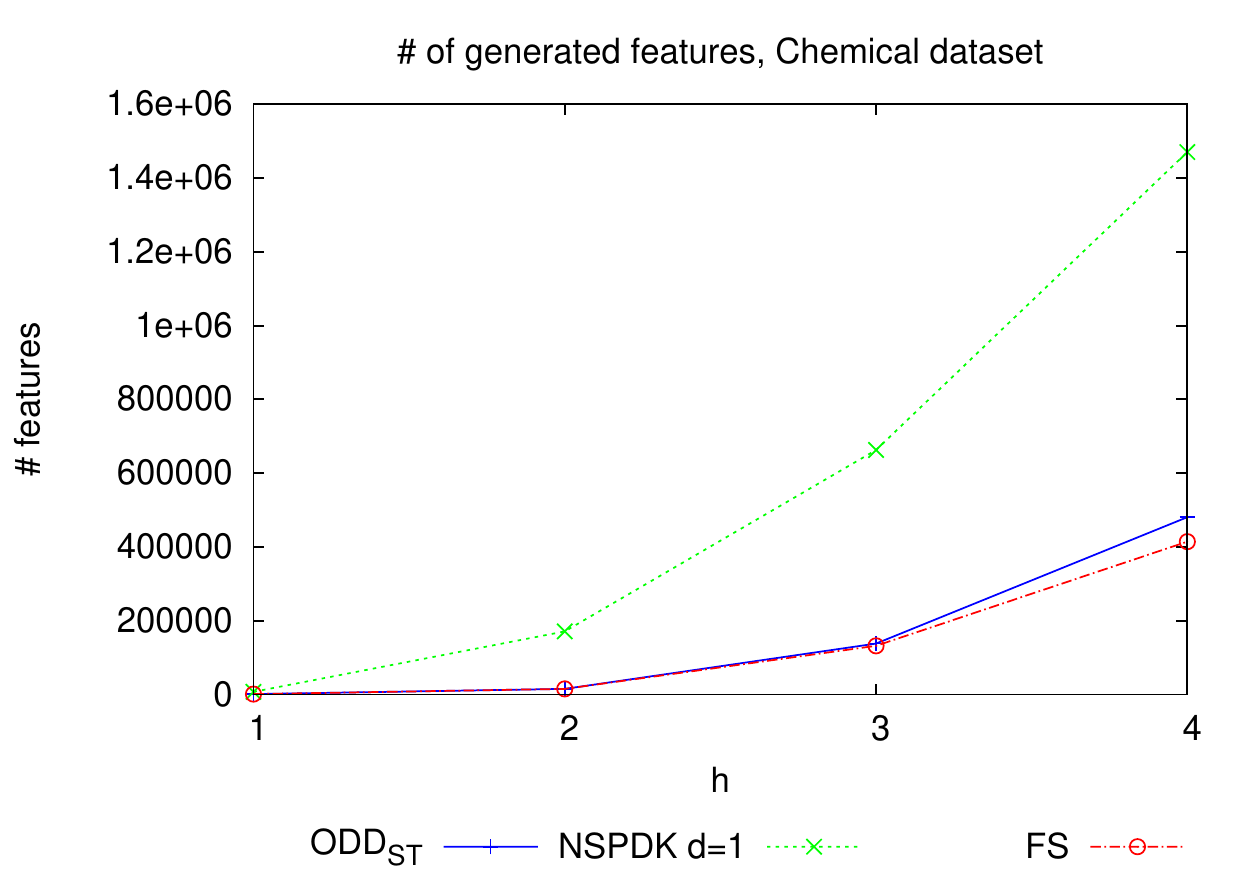}
 \caption{\label{fig:Chem2nfeatures} \rosso{Cumulative number of (different) features generated over the \textit{Chemical} stream according to the ODD$_{ST}$, NSPDK and FS kernels, for diferent $h$ parameter values.}
 }
\end{figure}

\rosso{Nonetheless, if we consider the size of the feature space induced by the kernels on a whole dataset, the number of different features that are generated may be very high.
Figure~\ref{fig:Chem2nfeatures} shows the size of the induced feature space for one of the datasets we will adopt in the experimental part of the paper (see Section~\ref{sec:results}), for different values of the $h$ parameter, for the considered kernels.
} 

\subsection{Dual Online Kernel Methods On a Budget} \label{sec:dualkernels}

The majority of online kernel methods on a budget are a variant of the perceptron \cite{Rosenblatt1958} and thus share a common structure. 
Let us assume the input stream is formed by pairs $e_t=(x_t,y_t)$, where $x_t\in\mathbb{X}$ is an input instance 
 and $y_t=\{-1,1\}$ is its label\footnote{As in the standard online setting, we assume that the target value $y_t$ is observed only after the system has predicted an output
 for $x_t$.}. 
 The goal is to find a hypothesis $h:\mathbb{X}\rightarrow \{-1,1\}$ such that the expected value of the adopted error measure on the stream is minimized. 
In the version of the perceptron we introduce here, which we call {\it Dual} since it is expressed in the kernel dual space (input space), the hypothesis is represented by a subset $M$ of the input instances \cite{Kivinen2004}. $M$ is commonly referred to as the model. 
The following is a general scheme of the {\it Dual} version of the perceptron: 
\begin{algorithm}[H]
\begin{algorithmic}[1]
	\scriptsize
    \STATE {\bfseries Input:} $\beta$ (algorithm dependent), $B$ (budget size)
    \STATE Initialize $M$: $M=\{\}$
    \FOR{{\bfseries each round} $t$}
      \STATE Receive an instance $x_t$ from the stream 
      \STATE Compute the score of $x_t$: $S(x_t)=\sum_{i=1}^{|M|} y_i\tau_i K(x_i, x_t)$
      \STATE Receive the correct classification of $x_t$: $y_t$
      \IF {$y_tS(x_t)\leq \beta$ ($x_t$ incorrectly classified)}
	\WHILE {$|M|+|x_t|>B$} \label{lst:line:modelsize}
	  \STATE select an element $x_j\in M$ for removal 
	  \STATE $M=M \setminus \{x_j\}$
	\ENDWHILE \label{lst:endwhile}
      \STATE update the hypothesis: $M=M\cup \{(y_t\tau_t,x_t)\}$
  \ENDIF
\ENDFOR
\end{algorithmic}
\caption{A general {\it Dual} perceptron-style algorithm for online kernel learning on a budget.\label{alg:implicit}}
\end{algorithm}
\noindent In Algorithm~\ref{alg:implicit}, $|M|$ represents the size of the model, i.e. the sum of the size of the instances in $M$. In the same way $|x_t|$ is the size of $x_t$. If the input instances are vectorial data, their size is constant, thus in order to add an element to $M$, it is sufficient to remove only one instance from $M$, i.e. the while loop in Algorithm~\ref{alg:implicit}\nero{-line 8} is executed exactly once. 
\nero{As it will be detailed in Section~\ref{sec:budgetalgorithms}, this is not the case in our scenario where the input instances are graphs and their size is not constant.} 
\rosso{Note that Algorithm~\ref{alg:implicit} tries to use as much memory as it is allowed to (without exceeding the limit $B$): line $8$ shows that one example would be removed from the model only if the algorithm, by inserting a novel example in the model, exceeded the memory limit $B$. In all other cases, any new erroneously-classified example is inserted in the model (line 12). All we shall see, the same observation will apply to the two other algorithms presented in this paper. } 

Many online algorithms can be seen as instances of Algorithm~\ref{alg:implicit}. 
For example, by setting $B=\infty$, $\tau=1$, $\beta=1$, we obtain the dual perceptron~\cite{Kivinen2004}. 
The Online Passive-Aggressive algorithm \cite{Crammer_Dekel_Keshet_Shalev-Shwartz_Singer_2006} tries to select an hypothesis with a unit margin on the examples. It is obtained with $B=\infty$, $\beta=1$,  $\tau_i=\min\left\{C,\frac{1-S(x_i)}{K(x_i,x_i)}\right\}$, where $C$ is a user-defined non-negative parameter. 
\mbox{In \cite{DBLP:journals/jmlr/OrabonaKC09}} it is described an update rule which tries to project the new instance onto the span of the current support set $M$. The resulting hypothesis is compared to the one obtained by inserting the whole instance into the model: if the difference between the two hypotheses is not greater than a user-defined threshold, then only the projected instance is added to the model. Computing the projection requires quadratic time and space with respect to the size of the support set, thus severely limiting the application to graph streams. 
Since the three algorithms assume $B=\infty$, no elements are removed from $M$. Thus, even if they try to minimize the size of the model, they do not provide any strategy to ensure that such size will not exceed any a priori given budget.  


When the problem setting imposes a budget $B$ on the size of the model, various strategies can be employed for selecting which elements should be removed from $M$. In \cite{Cavallanti:2007:TBH:1296038.1296052} the elements to be removed are chosen randomly. The Forgetron removes the oldest example in $M$ \cite{Dekel:2006}: a decay factor is applied to the $\tau$ values in such a way that the oldest examples in $M$ have lower and lower impact on the computation of $S()$. \mbox{Crammer et al. \cite{NIPS2003_AA29}} proposed to remove from $M$ any redundant example, i.e. the example with least impact on the margin of the hypothesis. This approach, however, is computationally expensive and thus it is not suitable for processing high dimensional data streams. 
In \cite{DBLP:journals/jmlr/WangV10} the Online Passive-Aggressive algorithm \cite{Crammer_Dekel_Keshet_Shalev-Shwartz_Singer_2006} has been extended to handle budget constraints. The idea is to modify the update rule such that the resulting hypothesis, after decreasing the model size such that the budget constraint is respected, has a small loss on the new example and it is similar to the current hypothesis. They describe three algorithms of increasing complexity and efficacy: BPA-S, BPA-NN, BPA-P. Among these, BPA-S has linear space and time complexity with respect to the model size. 

\subsection{Primal Algorithms for Online Learning On a Budget} \label{sec:primalalgs}

By the properties of kernel functions, each kernel evaluation corresponds to a dot product in an associated feature space.  
Then Algorithm~\ref{alg:implicit} has a corresponding version in feature space in which 
\nero{the examples are represented by their projection in feature space $\phi(x_t)\in \mathbb{R}^s$ (with $s$ being the size of the feature space). 
The hypothesis is represented by a vector $w\in \mathbb{R}^s$ \cite{Rosenblatt1958}, where the elements of $M$ are replaced by their sum: 
\begin{equation}
 w=\sum_{\phi(G_j)\in M} y_j\tau_j\phi(G_j). 
  \label{eq:w}
\end{equation}
}
The score is computed as $S(x_t)=w_t\cdot \phi(x_t)$ and the hypothesis is updated as $w_{t+1}=w_t+\tau_t y_t \phi(x_t)$. 
Given the fixed size of $w$, the standard perceptron does not take into account budget constraints. 
We refer to such version as {\it Primal}. 

An algorithm, similar to the one just described, has been presented in \cite{Langford:2009:SOL:1577069.1577097}: the update step is a stochastic gradient descent rule followed by a rounding step in which the small coefficients are set to zero. Since zero features may not be explicitly represented, the rounding phase allows to reduce the model size.
In \cite{Duchi:2009:EOB:1577069.1755882} a framework for minimizing a convex loss function together with a convex regularization term is presented. The update rule is constituted by two phases: the first one is a subgradient step with respect to the loss function and the second one looks for a vector which maximizes the similarity to the one obtained in the first phase while minimizing the regularization term. Various instantiations are discussed: among these, the one making use of the $\ell_1$ norm as a regularization term is interesting for this paper, since it promotes sparse solutions. 
Note that the literature on online learning algorithms working directly in feature space is incredibly vast, but here we are interested in algorithms corresponding to state of the art dual approaches. Indeed, our purpose is to assess the viability of primal approaches in the context of kernel methods. 

As for the algorithms discussed in Section~\ref{sec:dualkernels}, a drawback of the algorithms listed in this section is that, they do not provide any strategy to ensure that the size of the model $w$ will not exceed any a priori given budget.

\section{Budget-aware Algorithms for Structured Data} \label{sec:budgetalgorithms}

In this paper, we study three algorithms, together with different strategies for managing the budget, for graph streams. 
%
Our first proposal, Algorithm~\ref{alg:implicit}, needs a few adaptations before it can applied to graph data. 
Given the variable size of graph data we make use of the following measure for computing the size of the model in Algorithm~\ref{alg:implicit}: 
\begin{equation}
 |M|=\sum_{G_j\in M}( |V_{G_j}|+|E_{G_j}|+1), \label{eq:dualmodelsize}
\end{equation}
 where the constant term $1$ takes into account the occupancy of the value \nero{$\tau_ty_t$}. 
 The removal rule in Algorithm~\ref{alg:implicit} is modified as follows: when $G_t$ has to be inserted, instances are removed from $M$ until $|M|+ |V_{G_t}|+|E_{G_t}|+1 < B$, where $|M|$ is computed according to eq.~\eqref{eq:dualmodelsize}. 

The time complexity of an online algorithm depends on the number of graphs in $M$ and the complexity of the kernel function employed. 
In those settings in which the number of features associated with a kernel is not significantly greater than the size of the input, 
the evaluation of the kernel function may be greatly speeded up if it is performed as dot product of the corresponding feature vectors. 
Examples of kernels having such property are \cite{Shervashidze2009a, Costa2010, Dasan2012}. 
In the remainder of the section our observations will be restricted to this class of kernels. 
The actual size of  vectors $\phi(G)$ can be much less than $s$ if only non-null elements of $\phi(G)$ are represented in sparse format. 
We will refer to the number of non-null features of $\phi(G)$ as $|\phi(G)|$. 
These observations lead to the \textit{Primal/Dual} algorithm (referred to as {\it mixed} in the following):
\begin{algorithm}[H]
\begin{algorithmic}[1]
	\scriptsize
    \STATE {\bfseries Input:} $\beta$ (algorithm dependent), $B$ (budget size)
    \STATE Initialize $M$: $M=\{\}$
    \FOR{{\bfseries each round} $t$}
      \STATE Receive an instance $G_t$ from the stream 
      \STATE Compute the score of $G_t$: $S(G_t)=\sum_{\phi(G_j)\in M} y_j\tau_j\phi(G_j)\cdot \phi(G_t)$
      \STATE Receive the correct classification of $G_t$: $y_t$
      \IF {$y_tS(G_t)\leq \beta$ ($G_t$ incorrectly classified)}
	\STATE update the hypothesis:
	\WHILE{$1+\sigma|\phi(G_t)| + \sum_{\phi(G_j)\in M} 1+\sigma|\phi(G_j)|>B$}
	  \STATE select an element $\phi(G_j)\in M$ and remove it: $M=M\setminus\{\phi(G_j)\}$ 
	\ENDWHILE
    \STATE $M=M\cup \{y_t\tau_t\phi(G_t)\}$
  \ENDIF
\ENDFOR
\end{algorithmic}
\caption{{\it Mixed} perceptron-style algorithm for online learning on a budget.\label{alg:implicitexplicit}}
\end{algorithm}
Note that the model size is computed as $\sum_{\phi(G_j)\in M} 1+\sigma|\phi(G_j)|$, where the constant $1$ accounts for the \nero{$y_t\tau_t$} value and  $\sigma$ is the memory occupancy of a feature: if $\phi(G)$ is represented in sparse format as pairs ($i$, $\phi_i(G)$), where $\phi_i(G)$ is the value of the $i$-th feature of $G$, then $\sigma=2$. As we will see in Section~\ref{sec:implementation}, while $\sigma$ might be influenced by the budget management strategy employed, in all the experiments performed in this paper with Algorithm~\ref{alg:implicitexplicit} the value $\sigma$ will remain unchanged. 

Since in Algorithm~\ref{alg:implicitexplicit} the projection $\phi(G)$ is not computed for every kernel evaluation,  Algorithm~\ref{alg:implicitexplicit} is expected to be faster than Algorithm~\ref{alg:implicit}. However, if $|\phi(G)|>|V_{G_j}|+|E_{G_j}|$,  which generally holds, it uses more memory. 

Finally, we introduce  a budget online algorithm working in feature space. 
The idea is to replace all elements of $M$ with their sum as in eq.~\eqref{eq:w}. 
However, by so doing, we lose the connection between features and the instances they belong to. 
As a consequence, during the update of the hypothesis it is no more possible to select a whole vector  $\phi(G)$ for removal. Thus we propose to remove single features from $w$ when $|w|>B$ (here $|w|$ is the total number of non-null features appearing in any example added to the model). 
 
\begin{algorithm}[H]
\begin{algorithmic}[1]
    \scriptsize
    \STATE {\bfseries Input:} $\beta$ (algorithm dependent)
    \STATE Initialize $w$: $w_0=(0,\ldots,0)$
   \FOR{{\bfseries each round} $t$}
      \STATE Receive an instance $G_t$ from the stream 
      \STATE Compute the score of $G_t$: $S(G_t)= w_t\cdot \phi(G_t)$
      \STATE Receive the correct classification of $G_t$: $y_t$
      \IF {$y_t S(G_t)\leq \beta$ ($G_t$ incorrectly classified)}
	\WHILE{$\sigma|w+\phi(G_t)|>B$}
	  \STATE select a feature $i$ and remove it from $w$
	\ENDWHILE
	\STATE update the hypothesis: $w_{t+1}=w_t+\tau_t y_t \phi(G_t)$
      \ENDIF
   \ENDFOR
\end{algorithmic}
\caption{{\it Primal} perceptron-style online learning on a budget.\label{alg:explicitbudget}}
\end{algorithm}

The total memory occupancy of the model in Algorithm~\ref{alg:explicitbudget} reduces to $\sigma|w|$. 

Note that the elimination of the set $M$ allows Algorithm~\ref{alg:explicitbudget} to save a significant amount of memory while still being faster than Algorithms~\ref{alg:implicit} and \ref{alg:implicitexplicit}.

\subsection{Budget Management}\label{sec:implementation}

We have left unspecified how to select the examples/features to be removed when the budget is full in Algorithms~\ref{alg:implicit}-\ref{alg:explicitbudget}. 
\nero{As we briefly discussed in Section~\ref{sec:dualkernels}, complex strategies, which would require to solve an optimization problem, are usually expensive from the computational point of view~\cite{NIPS2003_AA29, DBLP:journals/jmlr/WangV10}. 
This is especially true for the graph domain for two main reasons. 
Graph data are generally high-dimensional thus making the solution of the optimization problems even more computationally expensive. 
The second reason is that, for instance the problem solved in \cite{DBLP:journals/jmlr/WangV10} (eq.~7) assumes that removing one example frees enough space for the novel example to be inserted, which does not hold for graphs since they are of variable size. Modifying the optimization problem to account for the removal of a subset of examples 
would increase the complexity of the problem, and the resulting method would not respect the constraint of linear processing time imposed by the setting considered in this paper.
For such reasons, we focus in this paper on heuristics for selecting the elements to be removed from the model. }
Given the differences in how the model is represented in the three algorithms, different strategies for pruning the model can be applied. 
We have explored the following strategies for Algorithms~\ref{alg:implicit} and \ref{alg:implicitexplicit}: 
\begin{itemize}
  \item ``random'', examples are removed randomly with uniform probability;  
  \item ``oldest'', the oldest examples are removed;
  \item ``$\tau$'', the examples with lowest $\tau$ values are removed. If more than one example has such $\tau$ value, the candidate is randomly selected. 
\end{itemize}
Note that the implementation of the three strategies does not increase the memory occupancy of the model. 

Since any kernel method using the kernel functions in \cite{Shervashidze2009a, Costa2010, Dasan2012} can be performed in the primal space, it is possible to apply {\it feature selection} techniques, i.e. deleting non-informative features, in order to reduce noise in the data and the size of the model \cite{Chen}. 
A typical approach is to compute a statistical measure for estimating the relevance of each feature with respect to the target concept, and 
 to discard the less-correlated features. 
Before describing the strategies for pruning the model for Algorithm~\ref{alg:explicitbudget}, we introduce an example of such measure, the F-score \cite{Chen}.
In the traditional batch scenario, the F-score of a feature $i$ is defined for binary  classification tasks as follows:
\begin{equation} 
F\!s(i) = \displaystyle \cfrac{(AVG_{i}^{+} - AVG_i)^2 + (AVG_{i}^{-} - AVG_i)^2} { \cfrac{\displaystyle\sum_{j \in Tr^{+}} (f^{j}_{i} - AVG_{i}^{+})^2}{ |Tr^{+}| - 1} + \cfrac{ \displaystyle\sum_{j \in Tr^{-}} (f^{j}_{i} - AVG^{-}_{i})^2} {|Tr^{-}| -1}     }
\label{eq:fscore}
\end{equation}
\noindent where $AVG_{i}$ is the average \nero{value} 
of feature $i$ in the dataset, $AVG_{i}^{+}$ ($AVG_{i}^{-}$) is the average \nero{value} 
of feature $i$ in positive (negative) examples, $|Tr^{+}|$ ($|Tr^{-}|$) is the number of positive (negative) examples and $f^{j}_{i}$ is the \nero{value} 
of feature $i$ in the $j^{th}$ example of the dataset. 
Features that get small values of F-score are not very informative with respect to the binary classification task \footnote{\nero{Even though F-score is known not to take into accout correlation between features, we select that measure for computational reasons.}}. 
Eq.~(\ref{eq:fscore}) cannot be applied as is to a stream since instances arrive one at time. As a minor contribution, we rewrite an incremental version of the F-score. 
Let ${\cal I}^+_t$  (${\cal I}^-_t$) be the set of positive (negative) instances which have been observed from the stream after having read $t$ instances, then the F-score $F\!s(i,t)$ can be rewritten by using the following quantities:
\[ n^+_t = | {\cal I}^+_t|,\ \ \ f^{+}_i(t) = \sum_{j\in {\cal I}^+_t} f^{j}_{i},\ \ \  f^{2,+}_i(t) = \sum_{j\in {\cal I}^+_t} (f^{j}_{i})^2\]
\[n^-_t = |{\cal I}^-_t|, \ \ \ f^{-}_i(t) = \sum_{j\in {\cal I}^-_t} f^{j}_{i}, \ \ \ f^{2,-}_i(t) = \sum_{j\in {\cal I}^-_t} (f^{j}_{i})^2.\]
In fact, we have:
\[AVG_{i,t}^{+} = \frac{f^{+}_i(t)}{n^+_t}, \ AVG_{i,t}^{-} = \frac{f^{-}_i(t)}{n^-_t}\]
\[ AVG_{i,t} = \frac{f^{+}_i(t)+f^{-}_i(t)}{n^+_t + n^-_t} \]
and
\begin{equation} 
F\!s(i,t) = \displaystyle \cfrac{(AVG_{i,t}^{+} - AVG_{i,t})^2 + (AVG_{i,t}^{-} - AVG_{i,t})^2} { D^+_t + D^-_t}
\label{eq:fscore-inc}
\end{equation}
where 
\[D^+_t = \frac{f^{2,+}_i(t) -2AVG_{i,t}^{+}f^{+}_i(t)  +n^{+}_t (AVG_{i,t}^{+})^2 }{ n^{+}_t - 1}, \]
\[D^-_t = \frac{f^{2,-}_i(t) -2AVG_{i,t}^{-}f^{-}_i(t)  +n^{-}_t (AVG_{i,t}^{-})^2 }{ n^{-}_t - 1}.\]
By defining $\delta^+(t+1) =1$ if the $(t+1)$th instance is positive; otherwise $\delta^+(t+1) =0$, and $\delta^-(t+1)=1-\delta^+(t+1)$, the quantities of interest can be updated incrementally as follows:
\[ n^+_{t+1} = n^+_{t}+\delta^+(t+1) ,\ \ \  f^{+}_i(t+1) = f^{+}_i(t) + \delta^+(t+1) f^{j}_{i},\]
\[ f^{2,+}_i(t+1) = f^{2,+}_i(t) + (\delta^+(t+1) f^{j}_{i})^2.\]
\[ n^-_{t+1} = n^-_{t}+\delta^-(t+1) ,\ \ \  f^{-}_i(t+1) = f^{-}_i(t) + \delta^-(t+1) f^{j}_{i},\]
\[ f^{2,-}_i(t+1) = f^{2,-}_i(t) + (\delta^-(t+1) f^{j}_{i})^2.\]
In order to incrementally compute the F-score, we need to keep track, for each feature $i$, of the following quantities: $f^{+}_i(t), f^{-}_i(t), f^{2,+}_i(t), f^{2,-}_i(t)$. 

We have explored the following strategies for Algorithm~\ref{alg:explicitbudget}: 
\begin{itemize}
  \item \textit{random} strategy: features are removed randomly with uniform probability.
   This strategy does not affect the size of the model, which is thus obtained setting $\sigma=2$ in Algorithm~\ref{alg:explicitbudget}.
  \item \textit{weight}: first, all the features of the example which are already present in the model, are inserted. This maximizes the information of the algorithm without increasing memory occupation. Next, for each feature left $f$ of the example, the feature of the model with lowest \nero{absolute} $w_i$ value (the weight associated with feature $f_i$), is selected.  Note that if all the features in the model have their $w_i$ higher than $f$, then $f$ is not inserted. The size of the model when this strategy is employed is obtained setting $\sigma=2$ in Algorithm~\ref{alg:explicitbudget}. 
    \item \nero{\textit{oldest} strategy: similar to the \textit{weight} strategy, but in this case we remove the least recently used feature. We need to associate to each feature the time in which that feature has been last inserted/modified.
The size of the model is obtained setting $\sigma=3$.}
  \item \textit{F-score}: it is similar to the \textit{weight} strategy, the only difference being that the $w_i$ value is replaced by the F-score, computed according to eq.~\eqref{eq:fscore}. By using the incremental version of the F-score, the correct size of the model is obtained by  setting $\sigma=5$ in Algorithm~\ref{alg:explicitbudget}, since we need to keep track of the index $i$ and the four valued neessary to incrementally update the F-score. 
%
\end{itemize}
{Note that the F-score strategy has no correspondence for \textit{Mixed} and \textit{Dual} algorithms. This strategy removes from the model the features with the lowest associated F-score. F-score measures the correlation of a feature with the target (+1 or -1). Indeed, a feature can appear in different examples, some positive and some negative. If there is a strong correlation with either class, the F-score of a feature will be high. 
On the contrary, \textit{Mixed} and \textit{Dual} algorithms remove whole examples from the budget. Since an example have a single label associated, that can be +1 or -1, it is not possible to compute correlation measures in this case.}\\

\section{Experimental results}\label{sec:experiments}

In this section, we empirically compare Algorithms~\ref{alg:implicit}-\ref{alg:explicitbudget} with state-of-the-art kernel functions for graphs described in Section~\ref{sec:oddkernel} and various budget management strategies on two graph datasets: the first one is composed of chemical compounds and the second one is composed of images. 
\nero{
Our purpose in this section is to study the performances, both in terms of prediction accuracy and running times, of the three algorithms as the memory budget varies, and to determine which algorithm is more appropriate for each setting.  
}

We start by describing in Section~\ref{sec:datasets} how the datasets were obtained. 
Then, in Section~\ref{sec:experimentalsetup}, we introduce the experimental setup and the adopted evaluation measure. 
Finally, the obtained results are illustrated and discussed in Section~\ref{sec:results}.

\subsection{Dataset Description} \label{sec:datasets}
\subsubsection{Chemical Dataset} \label{sec:chemicaldataset}

\begin{figure}
\centering
\begin{tikzpicture}[scale=1]
    \draw[->,yshift=-0.5cm,xshift=-5.4cm] (0,0) -- coordinate (x axis mid) (10.8,0);
                        \draw [yshift=-1cm,xshift=0cm]     node[anchor=north] {Number of graphs};
        \draw [yshift=-0.5cm,xshift=-3.0cm](8.4 cm,1pt) -- (8.4 cm,-3pt)
            node[anchor=north] {\hspace{-0.9cm}164,558};
                    \draw [yshift=-0.5cm,xshift=-3.75cm](3.75 cm,1pt) -- (3.75 cm,-3pt)
                   node[anchor=north] {82,279};

                    \draw [yshift=-0.5cm,xshift=-5.4cm](0 cm,1pt) -- (0 cm,-3pt)
            node[anchor=north] {0};
   \matrix[nodes={draw, ultra thick, fill=blue!20},
        row sep=0cm,column sep=0cm] {
     \node[rectangle] {\small AID:123 t=40}; &
          \node[rectangle] {\small AID:109  t=41}; &
                    \node[rectangle] {\small AID:123  t=47}; &
                    \node[rectangle] {\small AID:109  t=50};   \\
     };
     
   \end{tikzpicture}
   \caption{\label{fig:dataset2}Composition of the stream of graphs on chemical data. Four different target concepts are obtained by using different threshold values ($t$) on the activity scores of the compounding datasets.}
\end{figure}
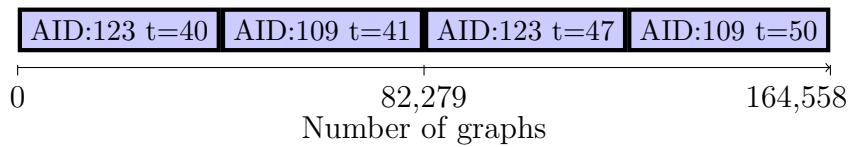

We have created graph streams combining two graph datasets available from the PubChem website (http://pubchem.ncbi.nlm.nih.gov).  
PubChem is a source of chemical structures of small organic molecules and their biological activities. 
It contains the bioassay records for anti-cancer screen tests with different cancer cell lines. Each dataset belongs to a certain type of cancer screen. For each compound an activity score is reported. 
The activity score for the selected datasets is based on increasing values of -LogGI50, where  GI50 is the concentration of the compound required for 50\% inhibition of tumor growth. 
A compound is classified as active (positive class) or inactive (negative class) if the activity score is, respectively, above or below a specified threshold. By varying the threshold we were able to simulate a drift on the target concept.  
 Our dataset is a combination of the ``AID: 123'' and ``AID: 109'' datasets from PubChem. 
 In ``AID:123'', growth inhibition of the MOLT-4 human Leukemia tumor cell line is measured as a screen for anti-cancer activity. The dataset comprises $40,876$ compounds, each one represented by a graph, tested at $5$ different concentrations. The average number of nodes for each graph in this dataset is $26.8$, while the average number of edges is $57.68$.
In ``AID:109'', growth inhibition of the OVCAR-8 human Ovarian tumor cell line is measured as a screen for anti-cancer activity on $41,403$ compounds. The average number of nodes for each compound is $26.77$, while the average number of edges is $57.63$. 
For each dataset, we used two different threshold values to simulate the concept drift: the median of the activity scores and the value such that  approximately $3/4$ of the compounds are considered dataset to be inactive (negative target). 
Finally, the stream has beeen obtained as the concatenation of ``AID: 123'' with threshold $1$, ``AID: 109'' with threshold $1$, ``AID: 123'' with threshold $2$, ``AID: 109'' with threshold $2$ (Figure~\ref{fig:dataset2}).
We call this stream \textit{Chemical}. 
Note that the stream is composed by four different concepts and comprises a total of $164,558$ graphs.
\rosso{Overall, the maximum number of nodes in a graph of the stream is 229, the maximum node outdegree is 6 and the alphabet size is 202.}
In order to assess the dependency of the results from the order of concatenation of the datasets, we created a second stream as:``AID: 123'' with threshold $1$, ``AID: 123'' with threshold $2$, ``AID: 109'' with threshold $1$, ``AID: 109'' with threshold $2$. Since the results were very similar to the ones obtained for the first dataset, for the sake of space, we do not report here the results for this second stream.
It should be stressed that the selected datasets represent very challenging classification tasks, independently of the value selected as the activity score threshold. 

\subsubsection{Image Dataset} \label{sec:imagedataset}

We created a stream of graphs from the \textit{LabelMe} dataset\footnote{http://labelme.csail.mit.edu/Release3.0/browserTools/php/dataset.php}. 
The dataset comprises a set of images whose objects are manually annotated via the LabelMe tool \cite{Russell2008}. 
The images are divided into several categories. We have removed those images having less than $3$ annotations. We have selected six categories amongst the ones having the largest number of images: ``office'' ($816$), ``home'' ($928$), ``houses'' ($1,294$), ``urban\_city'' ($865$), ``street'' ($1,069$), ``nature'' ($370$). In total we considered $5,342$ images. 

\begin{figure}
\centering
 \includegraphics[width=0.8\linewidth]{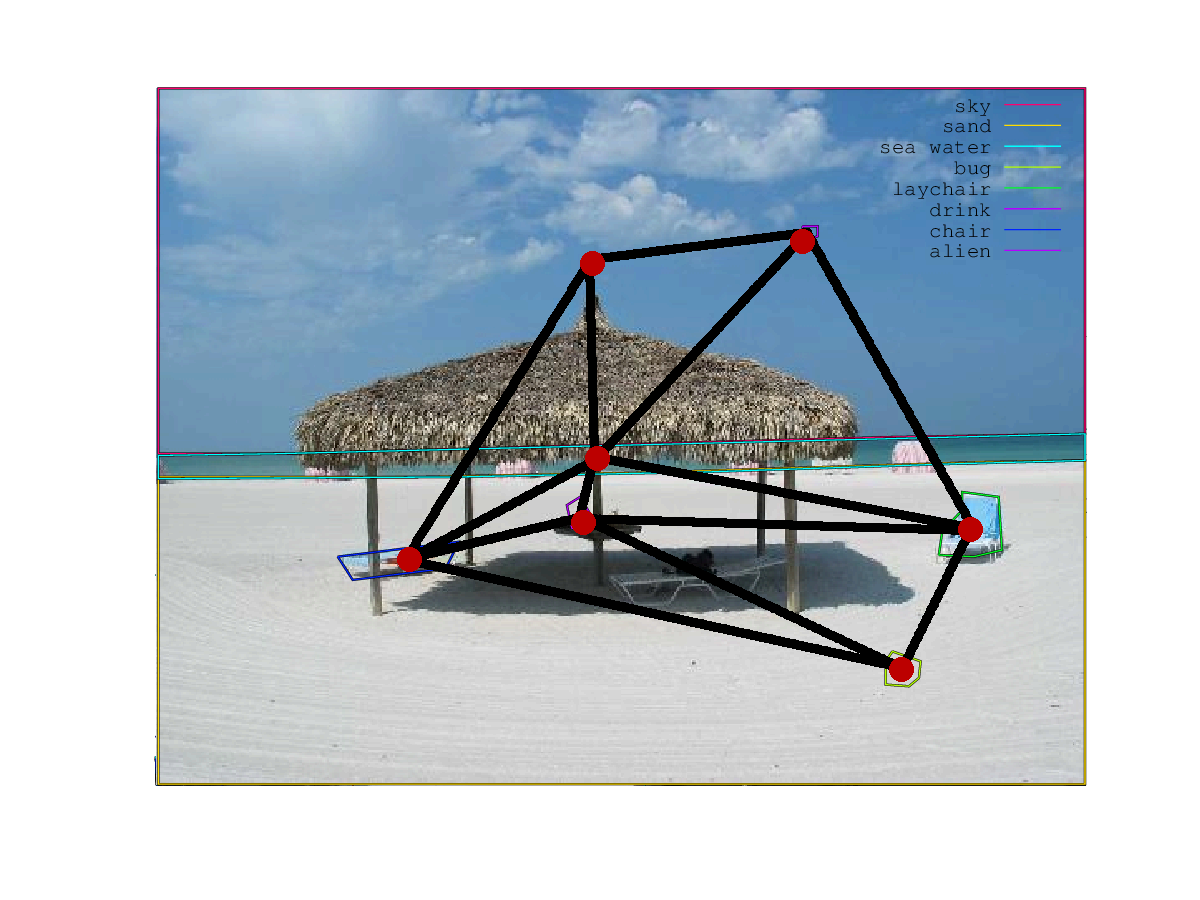}
 \caption{\label{fig:delaunay} An example of graph construction from an annotated image.}
\end{figure}

We then transformed each image into a graph: the annotated objects of the image become the nodes of the graph. The edges of the graph are determined according to the Delaunay triangulation \cite{delaunay}. The basic idea of the Delaunay triangulation is to connect spatially neighbouring nodes. Figure~\ref{fig:delaunay} gives an example of the construction of a graph from an image. 
The average number of nodes per graph is $14.37$ and the average number of edges is $63.61$. 

 The stream is made up of six parts \nero{(each part representing a different concept)}, for each of them one of the categories is selected as the positive class while the remaining ones represent the negative class; \nero{in order to simulate concept drifts} each one of the $5,342$ images appears six times in the stream: once with a positive class label, and $5$ times with negative class label. 
 The total number of examples composing the stream is $32,052$, \rosso{the maximum number of nodes of a graph is 201, the maximum node degree is 46 and the alphabet size is 65}. 


 
\subsection{Experimental setup} \label{sec:experimentalsetup}

For all the considered algorithms
(\textit{Primal, Mixed} and \textit{Dual}),
the $\beta$ and $\tau$ values were instantiated as $\beta=1$, \mbox{$\tau_i=\min\left\{C,\frac{1-S(x_i)}{K(x_i,x_i)}\right\}$}, as described in~\cite{DBLP:journals/jmlr/WangV10} for the BPA-S algorithm.
We chose BPA-S among the three BPA algorithms presented in~\cite{DBLP:journals/jmlr/WangV10}, because: \textit{i)} the results in the original paper show that, while being the fastest algorithm, the accuracy with respect to the other BPA versions does not degrade significantly; \textit{ii)} using (also) BPA-P or BPA-NN would have increased significantly the total time required for the experimentation.

The $C$ parameter has been tested in the \mbox{set $\{0.01$,  $0.1$, $1.0 \}$} for both \textit{Chemical} and \textit{Image} datasets.
\nuovo{By varying the C value, the results of the comparison between the three algorithms do not change. Therefore we report here only the results related to C=$0.01$.}
%
In order to increase the robustness of the results, the three algorithms have been tested with three different graph kernels: 
\begin{itemize}
  \item the \textit{Weisfeiler-Lehman subtree kernel} \nero{(FS)} \cite{Shervashidze2009a} with parameter values\\ \mbox{$h=\{0,1,2,3,4,5,6,7,8\}$};
  \item the \textit{Neighborhood subgraph pairwise distance kernel} \nero{(NSPDK)} \cite{Costa2010} with parameter values $h=\{1,2,3,4\}$, \mbox{$d=\{1,2,3,4,5,6\}$}. 
    \item the \textit{ODD$_{ST}$} kernel \cite{Dasan2012} with parameter values \mbox{$\lambda=\{0.8, 1, 1.2, 1.4, 1.6, 1.8\}$}, $h=\{1,2,3,4\}$;
\end{itemize}
All the proposed algorithms have the same upper bound $B$ on memory usage (budget), and the memory occupancy of the algorithms is calculated for \textit{Dual} as in eq.~\eqref{eq:dualmodelsize}, for {\it Mixed} as of line $9$ of Algorithm~\ref{alg:implicitexplicit} and for {\it Primal} as described in line $8$ of Algorithm~\ref{alg:explicitbudget} (note that the size of the model for {\it Primal} also depends on the budget management strategy). 
We experimented with budget values between $10,000$ and $50,000$ {\it memory units} (assuming each memory unit can store a floating point or integer number) \nero{ for the \textit{Chemical} dataset, and between $1,000$ and $100,000$ for the \textit{Image} dataset}. 
Higher values, for both datasets, were not tested since the time needed for the \textit{Dual} Algorithm to terminate became excessive (more than 48 hours). 

As for the strategies for managing the budget, we focused on the ``oldest'' and ``$\tau$'' ones for \textit{Dual}  and \textit{Mixed} algorithms. We focused on the ``oldest'' and ``weight'' strategies for \textit{Primal} algorithm \nero{(where we recall that ``weight'' is similar in spirit to ``$\tau$'' in the \textit{Primal} setting). Moreover, we considered also the ``F-score'' strategy for the \textit{Primal} algorithm.}
 
The random strategy has not been implemented because it tends to have worse \mbox{performances \cite{DBLP:journals/jmlr/WangV10}}. 


\rosso{The class distribution on the streams is unbalanced, therefore the 
Area Under the Receiver Operating Characteristic (AUROC) and the Balanced Accuracy~\cite{Chen} were adopted as performance measure. 
The AUROC measure is equal to the probability that a classifier will rank a randomly chosen positive instance higher than a randomly chosen negative one, thus it avoids inflated performance estimates on imbalanced datasets.}
\rosso{Since the results computed with Balanced Accuracy are very similar to the ones computed with the AUROC, we report only the latter, being the AUROC  more popular than the Balanced Accuracy.} 

\begin{figure}
\centering
 \includegraphics[width=0.8\linewidth]{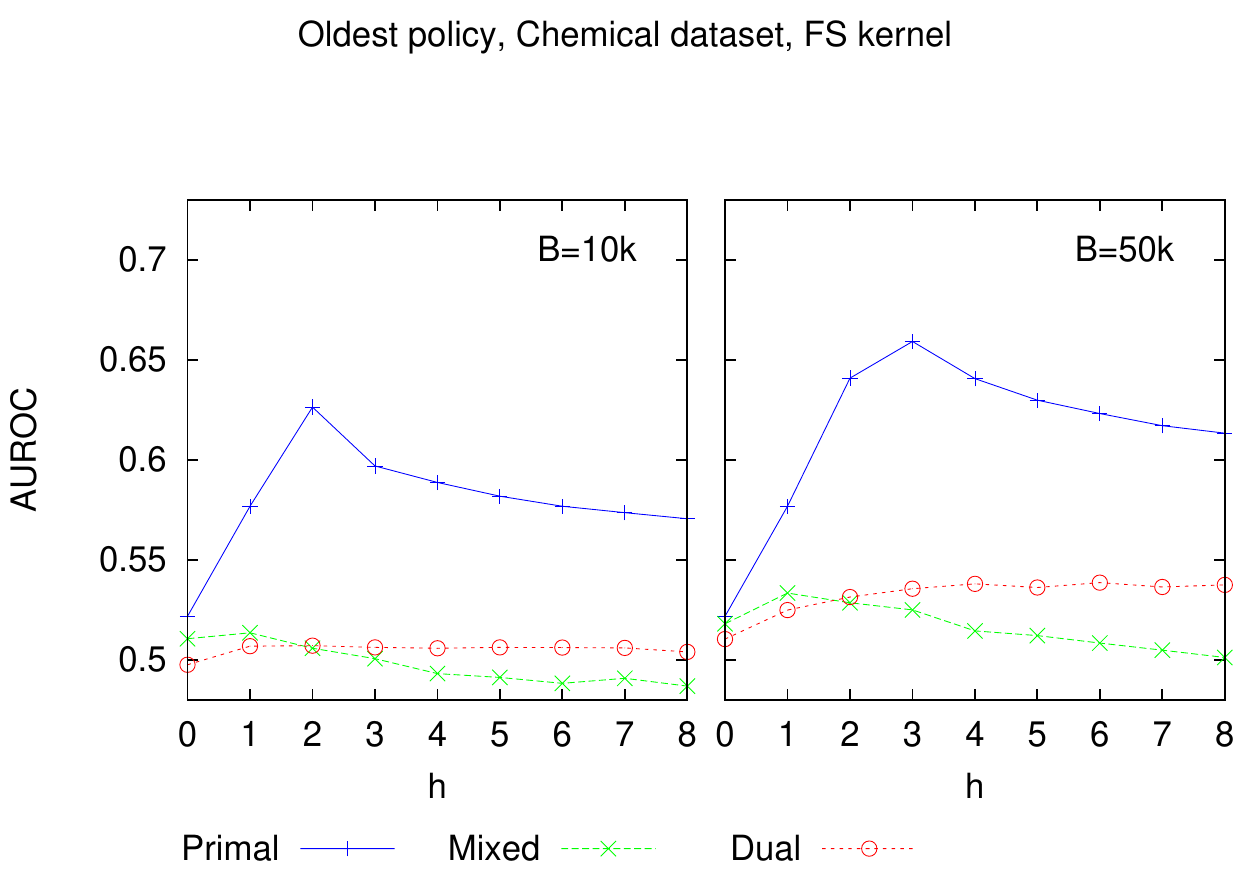}
  \includegraphics[width=0.8\linewidth]{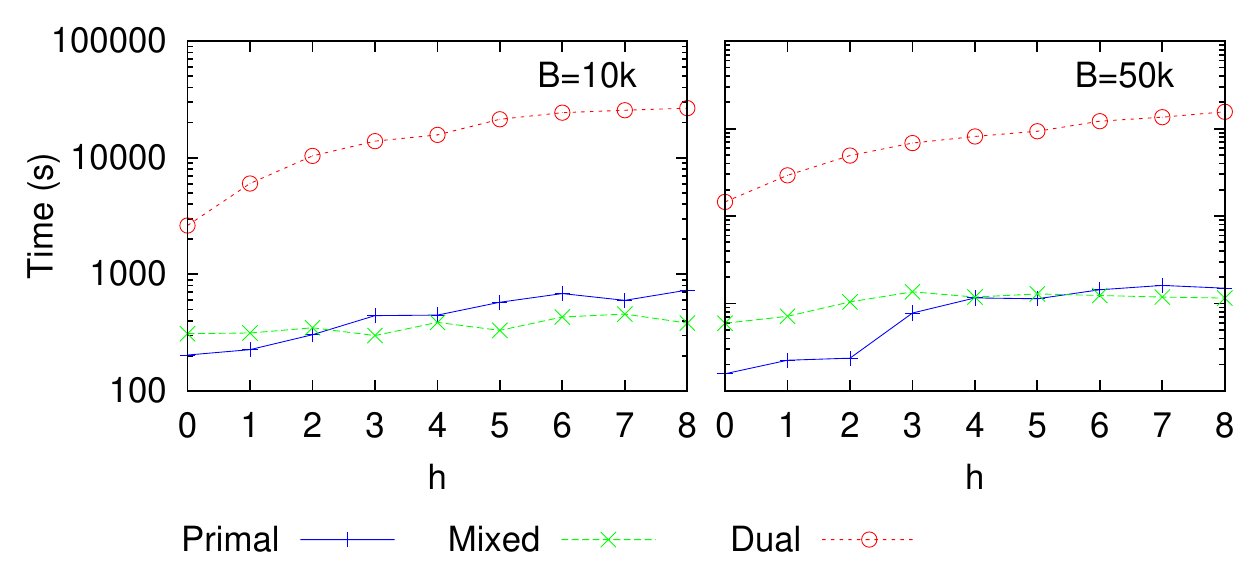}
 \caption{\label{fig:FSKChem2oldest} \rosso{Average AUROC value computed over all stream instances for memory budgets $B=10k$ (top left) and $B=50k$ (top right) for  algorithms \textit{Primal}, \textit{Mixed} and \textit{Dual} with respect to the values of the FS kernel parameter. Below each of the plots there is a second one with the corresponding running times}.  
 The plots refer to the \textit{Chemical} stream and the \textit{oldest} budget maintainance policy.
 }
\end{figure}

\begin{figure}
\centering
 \includegraphics[width=0.8\linewidth]{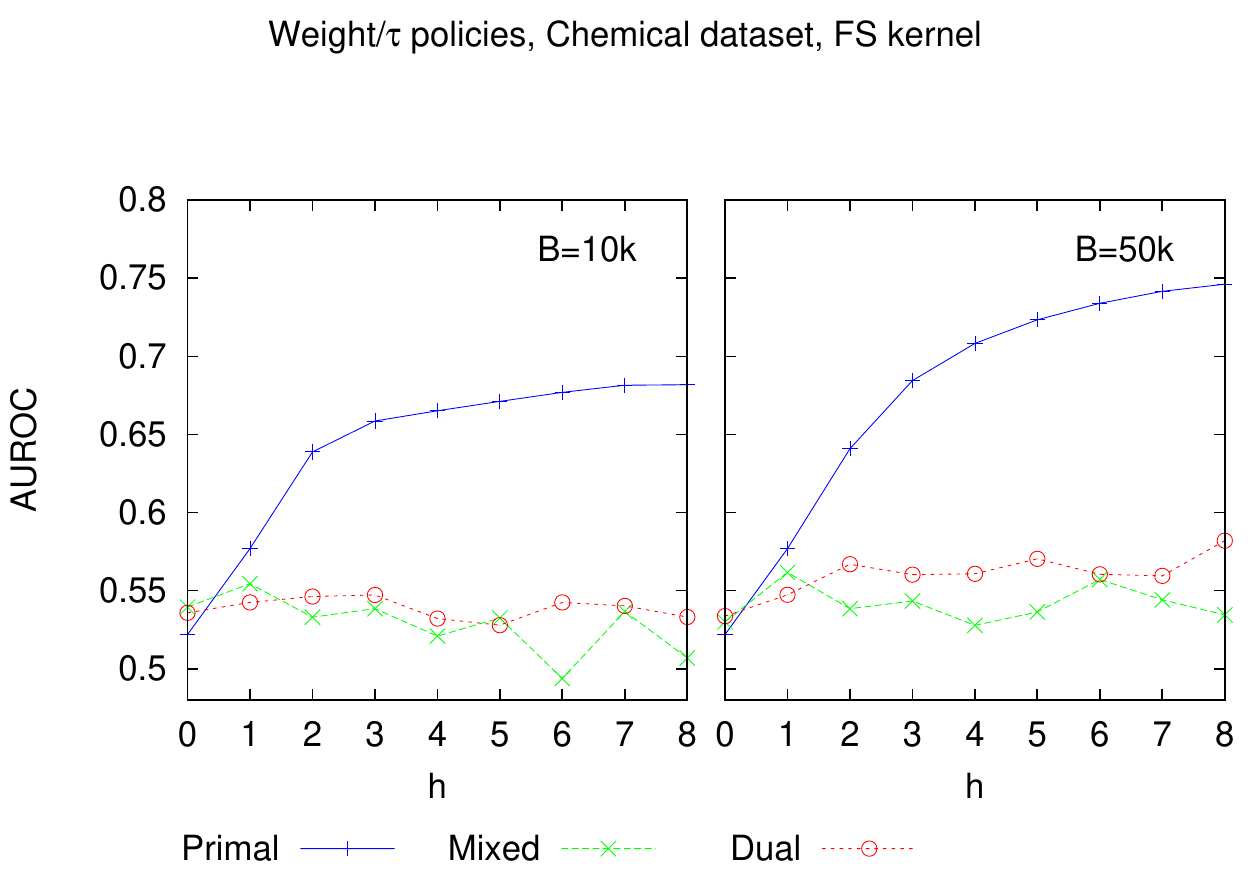}
   \includegraphics[width=0.8\linewidth]{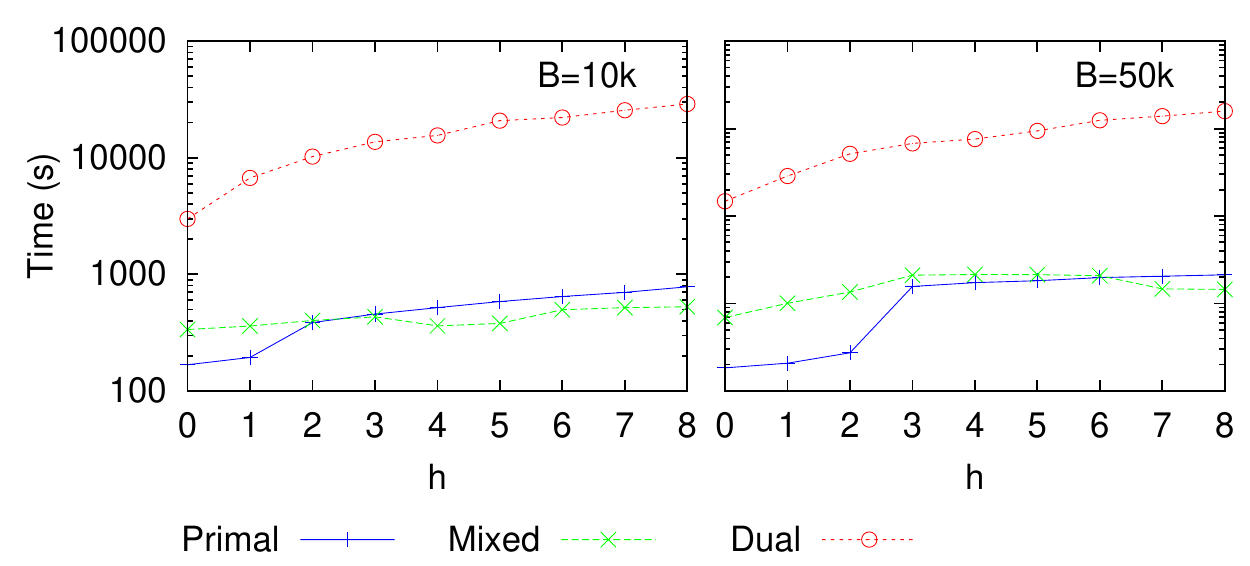}
 \caption{\label{fig:FSKChem2weight}  \rosso{Average AUROC value computed over all stream instances for memory budgets $B=10k$ (top left) and $B=50k$ (top right) for  algorithms \textit{Primal}, \textit{Mixed} and \textit{Dual} with respect to the values of the FS kernel parameter. Below each of the plots there is a second one with the corresponding running times}. 
 The plots refer to the \textit{Chemical} stream and the \textit{weight}/$\tau$ budget maintainance policies.}
\end{figure}

\begin{figure}
\centering
 \includegraphics[width=0.8\linewidth]{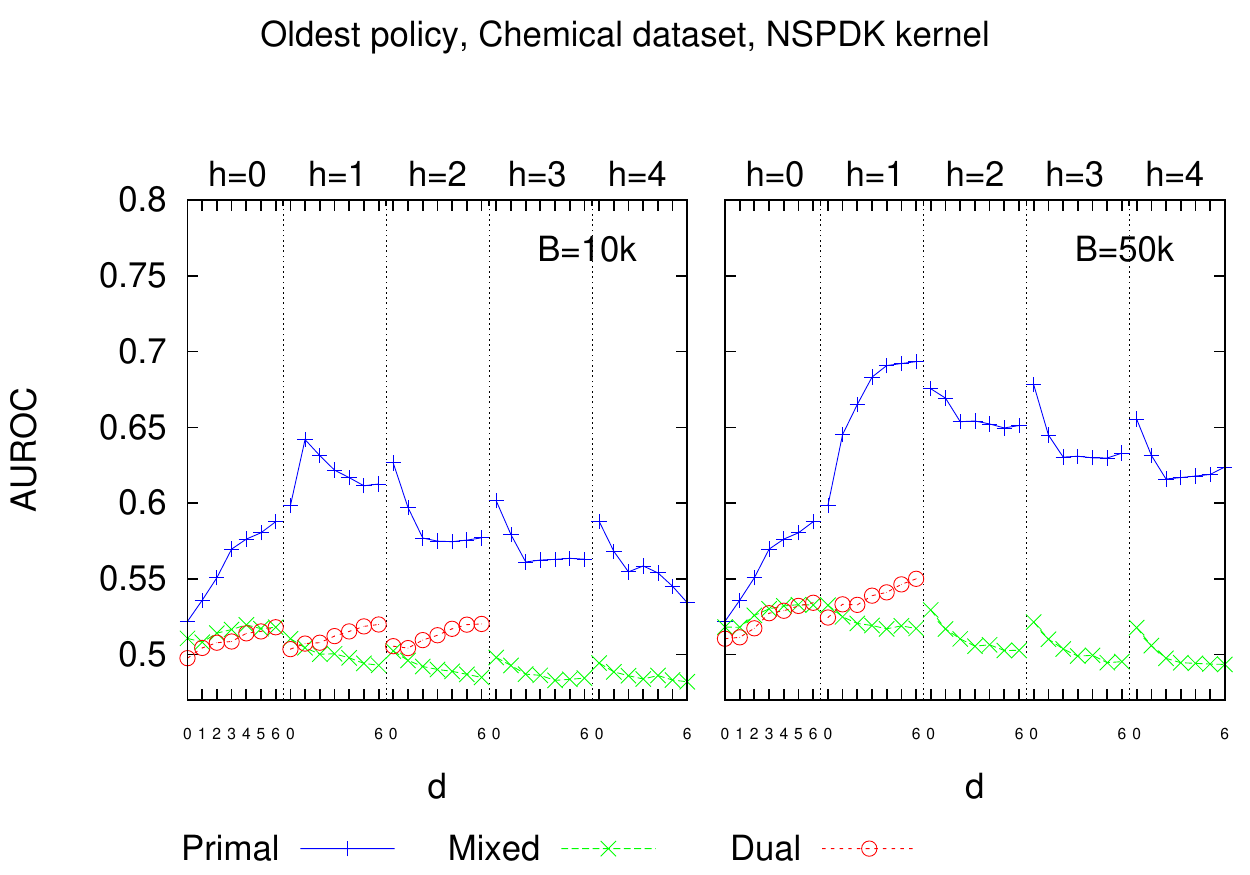}
    \includegraphics[width=0.8\linewidth]{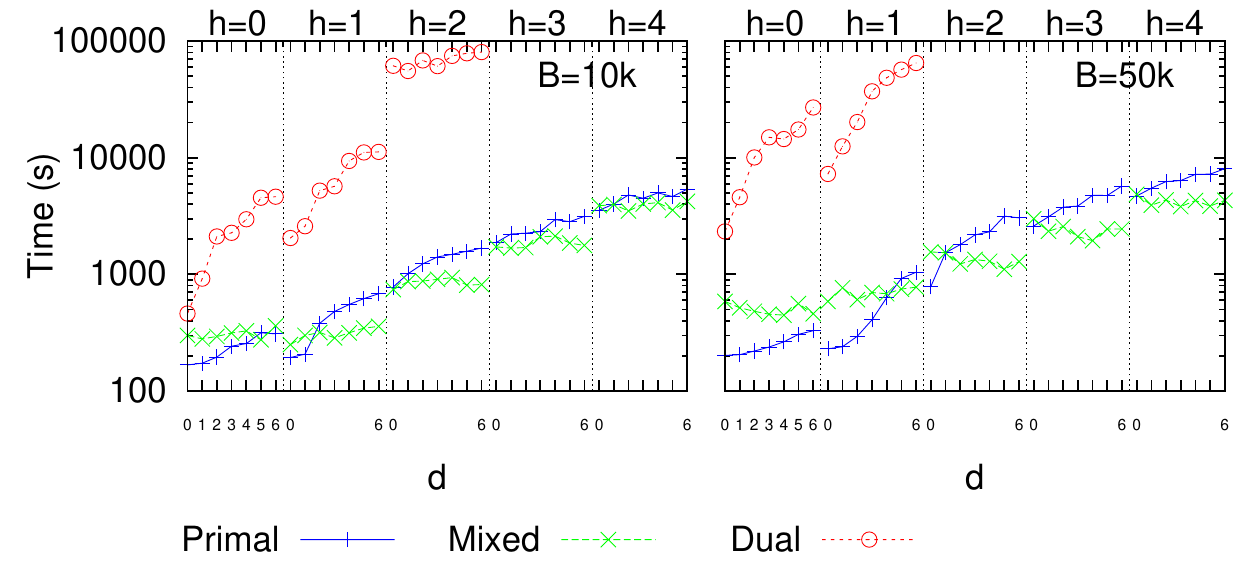}
 \caption{\label{fig:NSPDKChem2oldest} \rosso{Average AUROC value computed over all stream instances for memory budgets $B=10k$ (top left) and $B=50k$ (top right) for  algorithms \textit{Primal}, \textit{Mixed} and \textit{Dual} with respect to the values of the NSPDK kernel parameters. Below each of the plots there is a second one with the corresponding running times}. The plots refer to the \textit{Chemical} stream and the \textit{oldest} budget maintainance policy. \rosso{Missing values indicate that the corresponding execution has not terminated in 48 hours.} }
\end{figure}

\begin{figure}
\centering
 \includegraphics[width=0.8\linewidth]{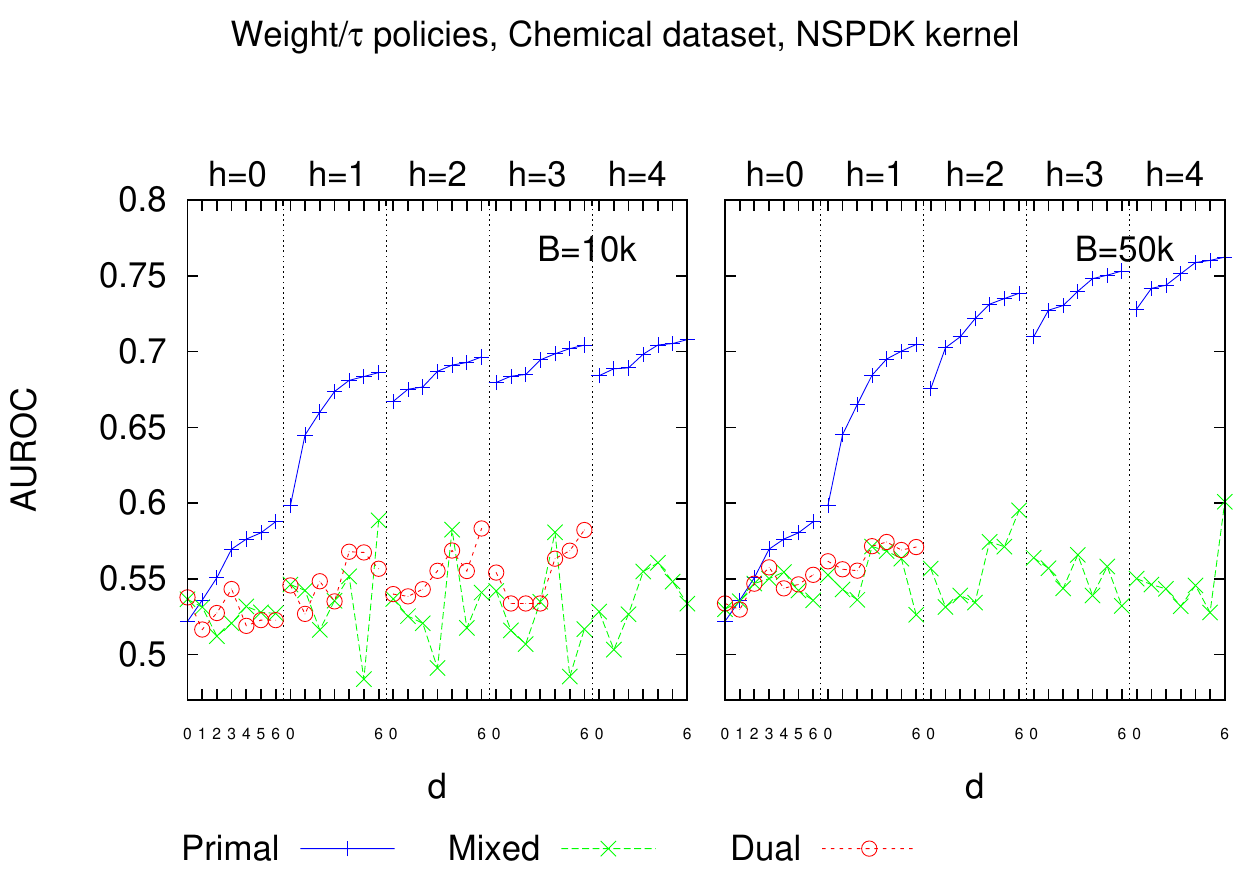}
     \includegraphics[width=0.8\linewidth]{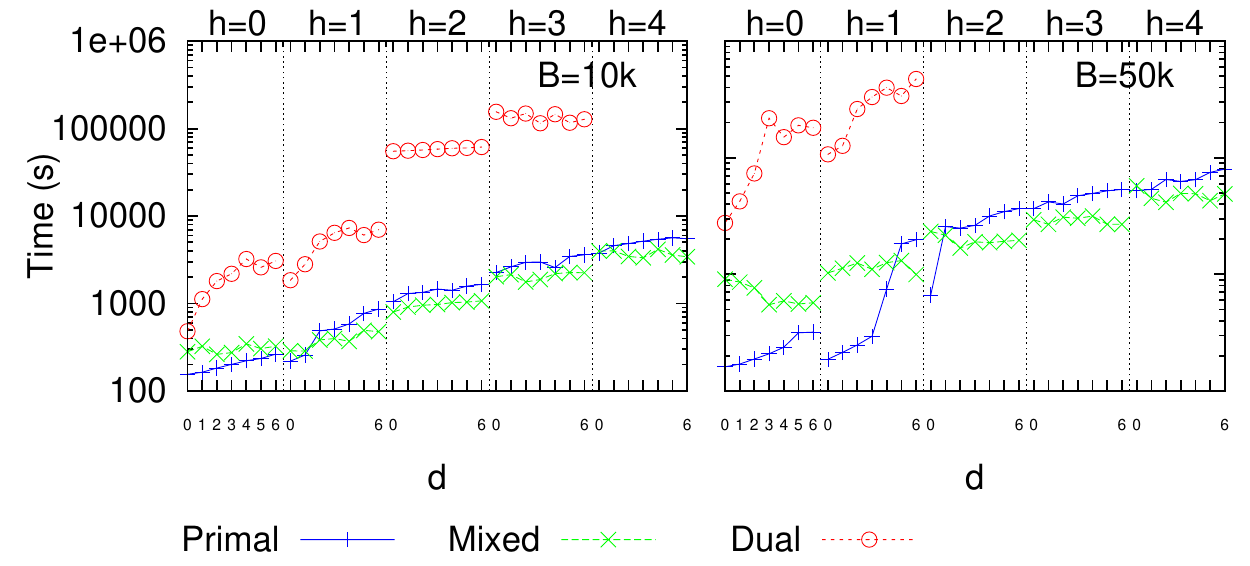}
 \caption{\label{fig:NSPDKChem2weight} \rosso{Average AUROC value computed over all stream instances for memory budgets $B=10k$ (top left) and $B=50k$ (top right) for  algorithms \textit{Primal}, \textit{Mixed} and \textit{Dual} with respect to the values of the NSPDK kernel parameters. Below each of the plots there is a second one with the corresponding running times}.The plots refer to the \textit{Chemical} stream and the \textit{weight}/$\tau$ budget maintainance policies. \rosso{Missing values indicate that the corresponding execution has not terminated in 48 hours.} }
\end{figure}

\begin{figure}
\centering
 \includegraphics[width=0.8\linewidth]{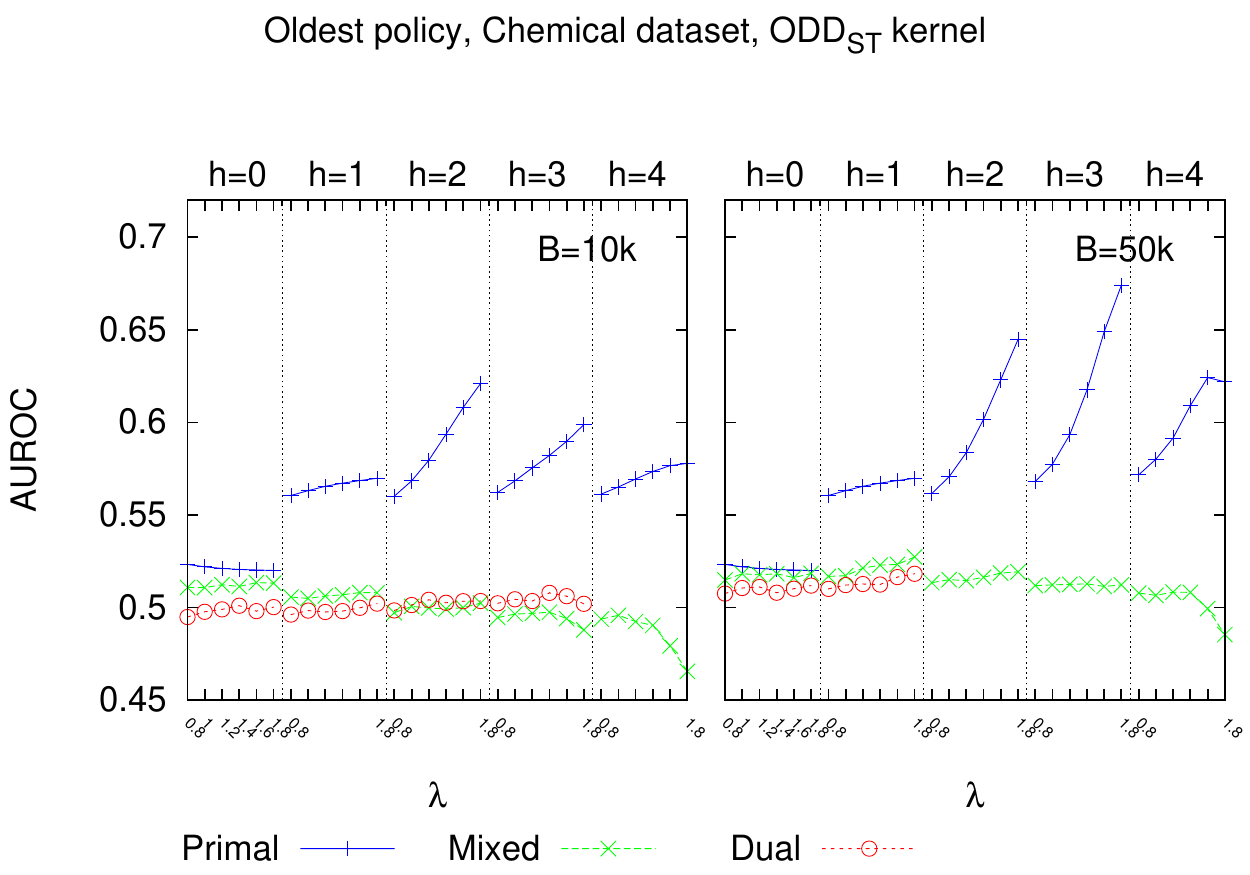}
     \includegraphics[width=0.8\linewidth]{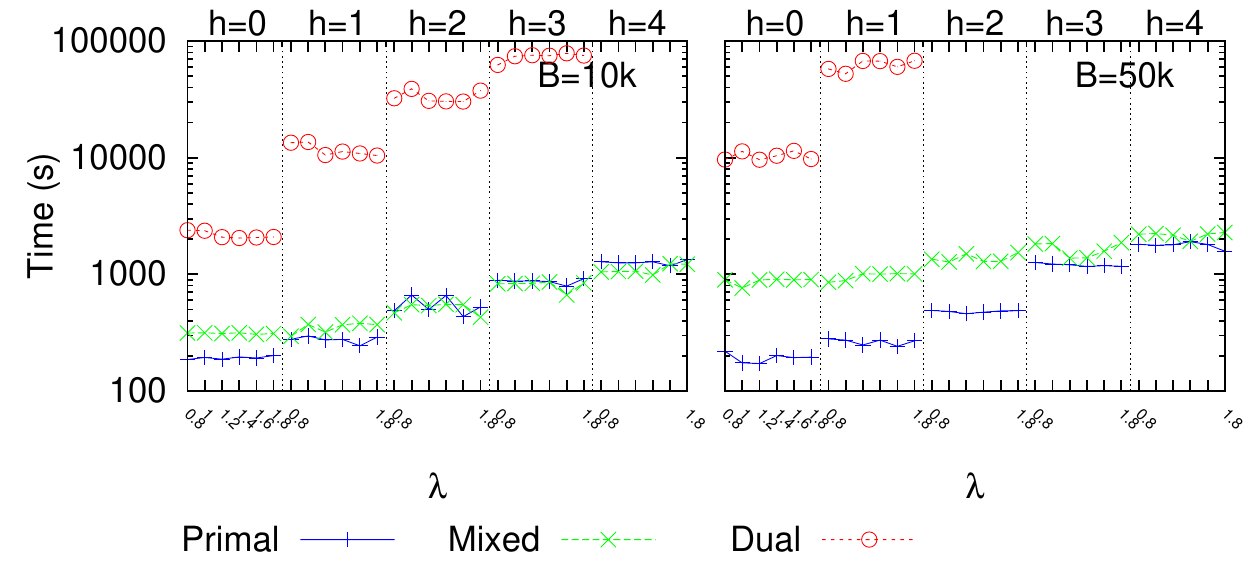}
 \caption{\label{fig:DDKChem2oldest} \rosso{Average AUROC value computed over all stream instances for memory budgets $B=10k$ (top left) and $B=50k$ (top right) for  algorithms \textit{Primal}, \textit{Mixed} and \textit{Dual} with respect to the values of the  $ODD_{ST}$ kernel parameters. Below each of the plots there is a second one with the corresponding running times}. The plots refer to the \textit{Chemical} stream and the \textit{oldest} budget maintainance policy. \rosso{Missing values indicate that the corresponding execution has not terminated in 48 hours.} }
\end{figure}

\begin{figure}
\centering
 \includegraphics[width=0.8\linewidth]{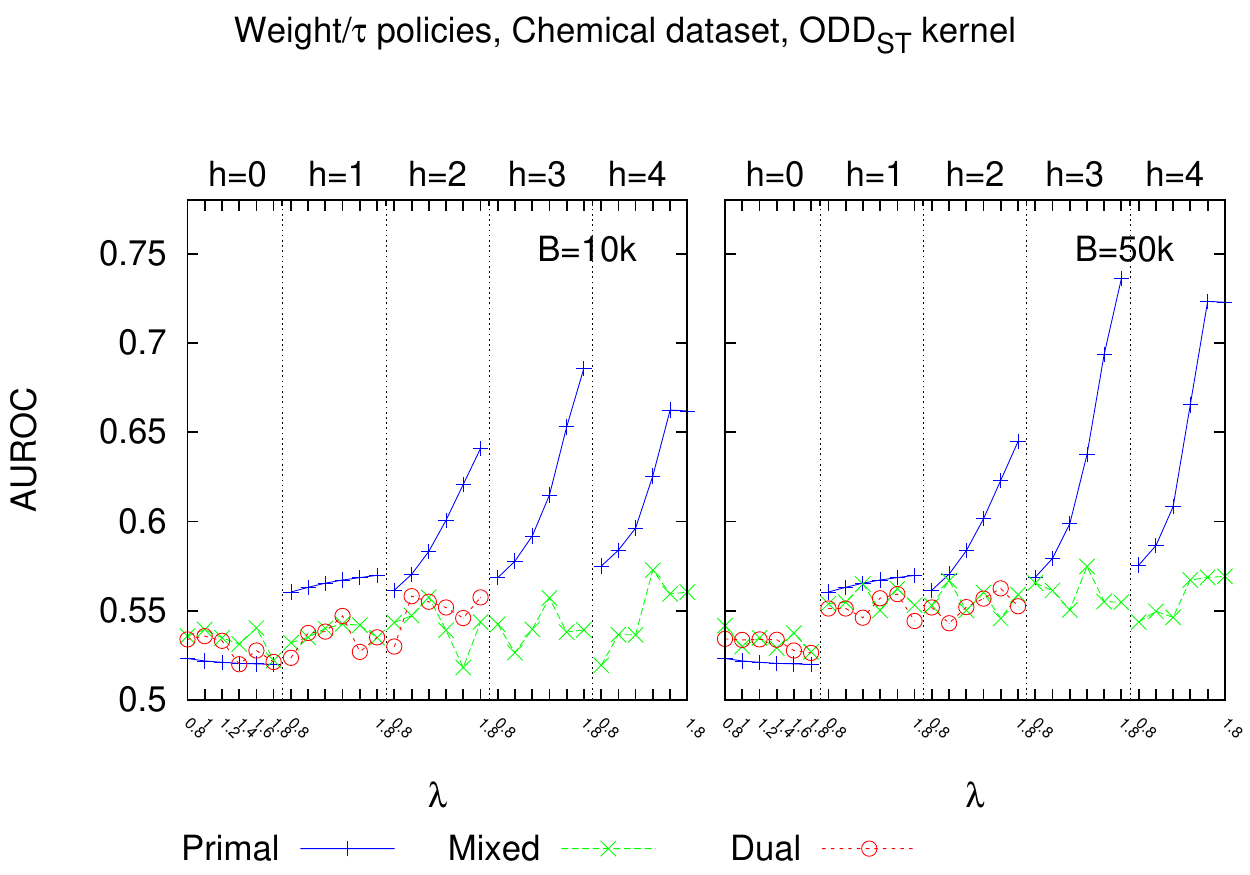}
      \includegraphics[width=0.8\linewidth]{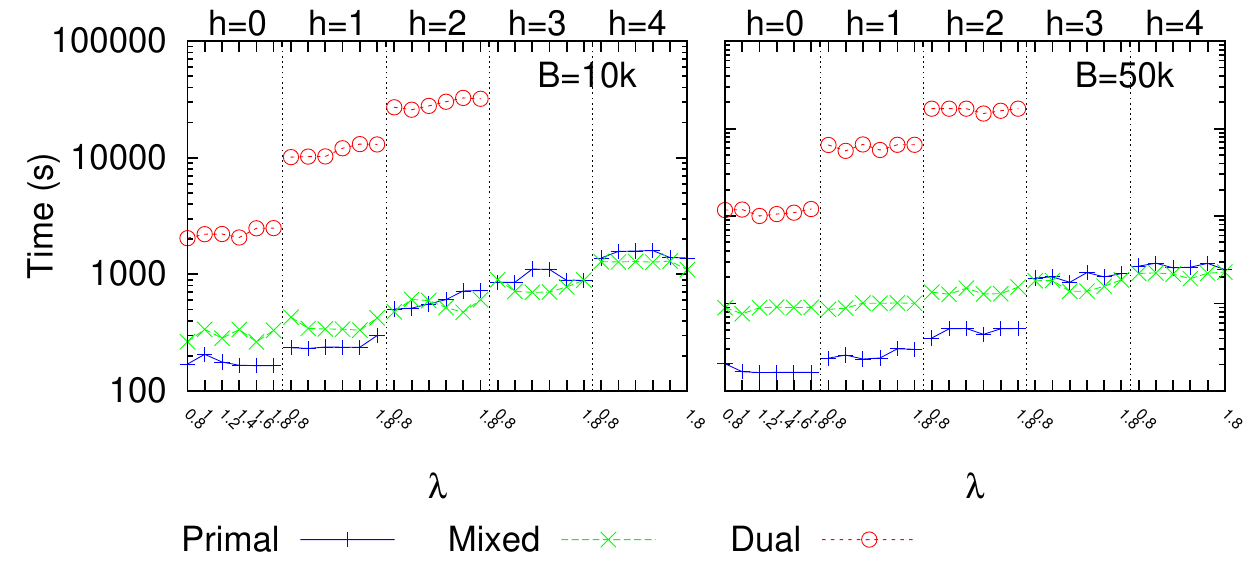}
 \caption{\label{fig:DDKChem2weight} \rosso{Average AUROC value computed over all stream instances for memory budgets $B=10k$ (top left) and $B=50k$ (top right) for  algorithms \textit{Primal}, \textit{Mixed} and \textit{Dual} with respect to the values of the  $ODD_{ST}$ kernel parameters. Below each of the plots there is a second one with the corresponding running times}. The plots refer to the \textit{Chemical} stream and the \textit{weight}/$\tau$ budget maintainance policies. \rosso{Missing values indicate that the corresponding execution has not terminated in 48 hours.}}
\end{figure}

The plots in Figures~\ref{fig:FSKChem2oldest}-\ref{fig:DDKChem2weight}, Figures~\ref{fig:FSKImageoldest}-\ref{fig:DDKImageweight} and Table~\ref{tab:results} regarding the AUROC measure are obtained as follows: for each run (Dataset/Kernel/parameters combination) the AUROC measure is sampled every $50$ examples. Then we compute the average over all samples and obtain a single value.
We chose not to show the behavior of each algorithm during a single run because we have performed more than 300 runs. 
 The running times are computed on a machine with two Intel(R) Xeon(R) CPU E5-4640@ 2.40GHz equipped with 256GB of RAM. Notice that the executions use a single core and a very limited amount of RAM. 

\subsection{Results and discussion} \label{sec:results}
\begin{figure}
\centering
 \includegraphics[width=0.85\linewidth]{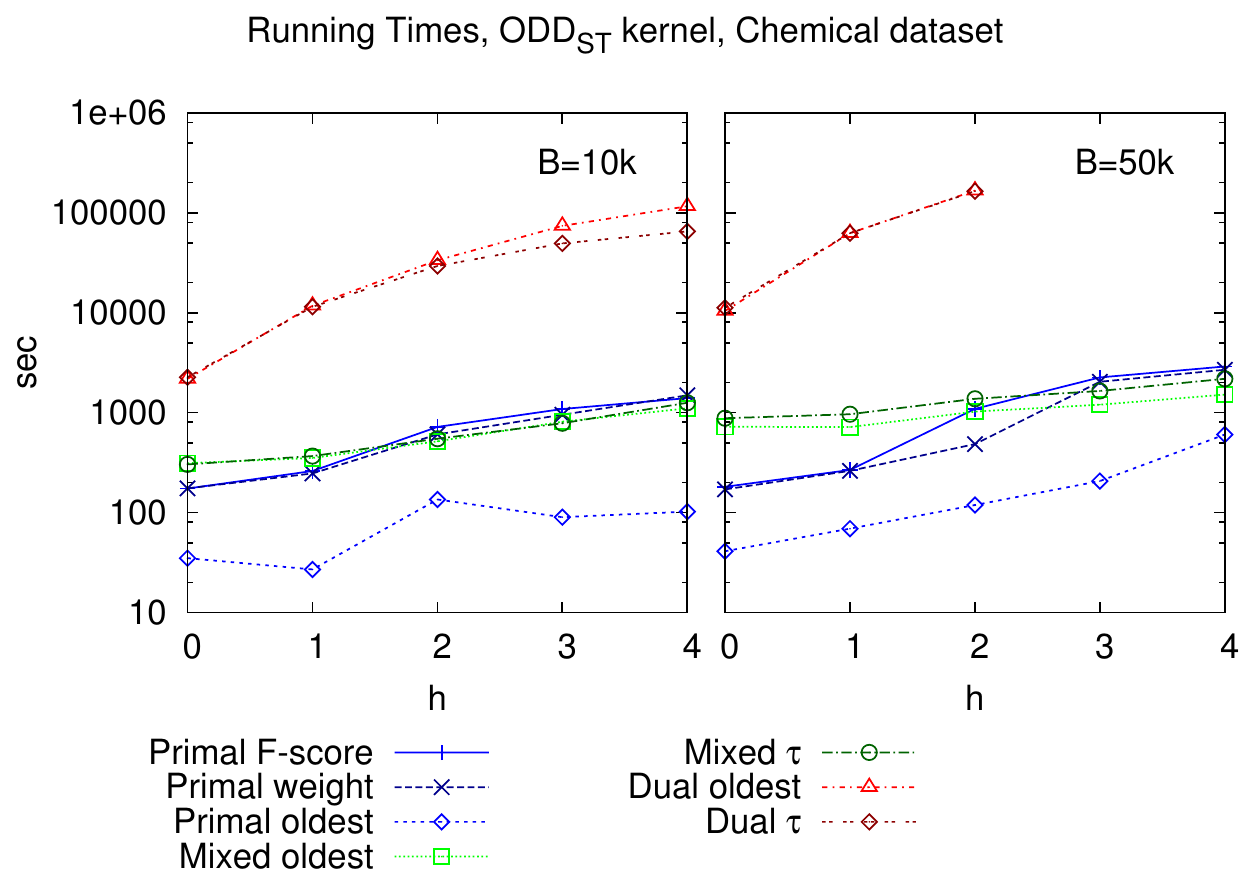}
 \caption{\label{fig:TimesDDKChem} Average computational times of algorithms {\textit Primal}, {\textit Mixed} and {\textit Dual} on the \textit{Chemical} dataset for the $ODD_{ST}$ kernel. }
\end{figure}
%
%

\nero{The aim of the experiments is to compare correspondent budget management strategies for \textit{Primal}, \textit{Dual} and \textit{Mixed}: \textit{i)} \textit{oldest} for the three algorithms; \textit{ii)} \textit{weight} for \textit{Primal} and $\tau$ for \textit{Mixed} and \textit{Dual}. 
For each of the above correspondent budget strategies, we observe the performances of the three algorithms, for each combination of kernel function and kernel parameters, as the budget varies.}
\rosso{Section~\ref{sec:expschemical} reports the experiments on the Chemical Dataset.
Section~\ref{sec:expsimage} reports the experiments on the Image Dataset.} 
Finally, section~\ref{sec:expsdiscussion} draws general conclusions on the experiments. 

\subsubsection{Experiments on the Chemical Dataset} \label{sec:expschemical}

\rosso{The Figures~\ref{fig:FSKChem2oldest}-\ref{fig:TimesDDKChem} report the results for one kernel, one specific budget management policy, two budget values, $B=10k$ and $B=50k$.
 Each Figure is divided into 4 subfigures: the ones on the left side refer to budget $B=10k$, the ones on the right refer to budget $B=50k$; the two figures on top report the AUROC measure, while the two on the bottom report running times.  
One point in a plot represents the AUROC/running time over all \textit{Chemical} dataset for one configuration of the kernel parameters. 
Note that running times are in logarithmic scale. } 

\nero{Figures~\ref{fig:FSKChem2oldest}-\ref{fig:FSKChem2weight} refer to the FS kernel with \textit{oldest} and \textit{weight} budget management policy, respectively. 
\rosso{Note that, by increasing $h$, the representation in memory of an example does not change for Algorithm~\ref{alg:implicit}, whilst it requires more memory for Algorithms~\ref{alg:implicitexplicit}-\ref{alg:explicitbudget} since the number of features increases.}  
 Figures~\ref{fig:NSPDKChem2oldest}-\ref{fig:NSPDKChem2weight} refer to the NSPDK kernel (with the same budget values). Each point refers to a combination of the $h$ and $d$ parameters of the kernel (the values are grouped with respect to the $h$ parameter).
 Figures~\ref{fig:DDKChem2oldest}-\ref{fig:DDKChem2weight} are similar but show the results referring to the $ODD_{ST}$ kernel (values are again grouped with respect to the $h$ parameter).
 Consider that, on this dataset and with no memory budget constraint on the model, the ODD$_{ST}$ kernel generates a model with a total of $91,467$ features with $h=3$ (the higher the $h$ parameter, the more features are generated). Such number is the size of w  ($||w||$) and thus the size of the vectorial representation of the model.
}
 \begin{table}[ht]
 \caption{Best \textit{AUROC} value ($\pm$ standard deviation) for each dataset, algorithm, policy, kernel for $10k$ and $50k$ budget values. \label{tab:results}}
 \centering
 \begin{tabular}{|c|c|c|c|c|c|c|}
 \hline
 kernel & Alg. & Policy & \multicolumn{2}{|c|}{Chemical} & \multicolumn{2}{|c|}{Image} \\
 & & & 10k & 50k &10k & 50k  \\
 \hline
 \multirow{6}{*}{\rotatebox[origin=c]{90}{FS} } & \multirow{3}{*}{Primal} &weight &\textbf{.681} \tiny{$\pm .094$} & \textbf{.746} \tiny{$\pm .096$  }&.914 \tiny{$\pm .094$ }& .913 \tiny{$\pm .095$} \\
    &  & oldest &  .626 \tiny{$\pm .092$} &.659 \tiny{$\pm .093$ }&\textbf{ .917} \tiny{$\pm .092$ }&.918 \tiny{$\pm .090$}\\
   &  & F-score&.644 \tiny{$\pm .096$}&  .669 \tiny{$\pm .096$}& .916 \tiny{$\pm .090$}& \textbf{.919} \tiny{$\pm .091$}\\
    & \multirow{2}{*}{Mixed} &$\tau$ &.554 \tiny{$\pm .124$ }& .561 \tiny{$\pm .114$} &.908 \tiny{$\pm .099$} &.901 \tiny{$\pm .095$} \\
   &  & oldest& .513 \tiny{$\pm .096$}& .533 \tiny{$\pm .097$}& .907 \tiny{$\pm .103$} &.912 \tiny{$\pm .096$} \\
     & \multirow{2}{*}{Dual} &$\tau$ &.547 \tiny{$\pm .127$ }& .582 \tiny{$\pm .115$} &.907 \tiny{$\pm .093$} &.906 \tiny{$\pm .094$} \\
   &  & oldest& .507 \tiny{$\pm .098$}& .538 \tiny{$\pm .098$} & .884 \tiny{$\pm .117$} &.915 \tiny{$\pm .090$} \\
   \hline
    \multirow{6}{*}{\rotatebox[origin=c]{90}{ NSPDK} } & \multirow{3}{*}{Primal} &weight &\textbf{.707} \tiny{$\pm .091$ }&\textbf{.762} \tiny{$\pm.092$} &.907 \tiny{$\pm .095$} &.907 \tiny{$\pm .095$} \\
        &  & oldest & .641 \tiny{$\pm .092$} & .693 \tiny{$\pm .092$} &.909  \tiny{$\pm .093$}& .910 \tiny{$\pm .092$}\\
   &  & F-score& .674 \tiny{$\pm .092$}& .691 \tiny{$\pm .090$} & \textbf{.914} \tiny{$\pm .091$} & .912 \tiny{$\pm .094$}\\
    & \multirow{2}{*}{Mixed} &$\tau$ & .588 \tiny{$\pm .126$}& .600 \tiny{$\pm .114$} & .894 \tiny{$\pm .100$} & .882 \tiny{$\pm .113$}\\
   &  & oldest& .519 \tiny{$\pm .101$}& .532 \tiny{$\pm .102$} & .899 \tiny{$\pm .106$} & .907 \tiny{$\pm .091$}\\
     & \multirow{2}{*}{Dual} &$\tau$ & .583 \tiny{$\pm .121$}& .581 \tiny{$\pm .103$} & .892 \tiny{$\pm .105$} & .890 \tiny{$\pm .105$}\\
   &  & oldest& .520 \tiny{$\pm .102$}& .571 \tiny{$\pm .083$} & .877 \tiny{$\pm .115$} & \textbf{.918} \tiny{$\pm .093$}\\
   \hline
    \multirow{6}{*}{ \rotatebox[origin=c]{90}{ODD$_{ST}$} } & \multirow{3}{*}{Primal} &weight & \textbf{.685} \tiny{$\pm .094$} & \textbf{.735} \tiny{$\pm .097$ }&\textbf{.919} \tiny{$\pm .088$}& \textbf{.919} \tiny{$\pm .088$}\\
    &  & oldest & .620 \tiny{$\pm .092$} & .674 \tiny{$\pm .094$} &\textbf{.919} \tiny{$\pm .088$} &\textbf{.919} \tiny{$\pm .088$}\\
   &  & F-score& .661 \tiny{$\pm .098$} & .693 \tiny{$\pm .097$} & \textbf{.919} \tiny{$\pm .088$}& \textbf{.919} \tiny{$\pm .088$}\\
    & \multirow{2}{*}{Mixed} &$\tau$ & .572 \tiny{$\pm .125$} & .574 \tiny{$\pm .125$} &.909 \tiny{$\pm .093$} & .905 \tiny{$\pm .107$} \\
   &  & oldest& .513 \tiny{$\pm .098$} & .527 \tiny{$\pm .095$} &.910 \tiny{$\pm .098$} & .917 \tiny{$\pm .085$}\\
     & \multirow{2}{*}{Dual} &$\tau$ & .558 \tiny{$\pm .134$} & .562 \tiny{$\pm .129$} &.907 \tiny{$\pm .096$}& .910 \tiny{$\pm .095$} \\
   &  & oldest& .504 \tiny{$\pm .097$} & .518 \tiny{$\pm .097$} &.883 \tiny{$\pm .120$}& .907 \tiny{$\pm .098$} \\
 \hline
 \end{tabular}
\end{table}

Table~\ref{tab:results} reports, for each combination of dataset, algorithm, kernel, policy and budget \nero{values $10k$ and $50k$}, the best AUROC value among the tested parameters. 
\rosso{The table allows to easily compare different policies and different algorithms.} 

If we consider the \textit{Chemical} dataset, the highest value for \textit{Primal} algorithm is $0.762$ (NSPDK, \textit{weight} policy, budget $50k$), while the best AUROC value for Algorithm \textit{Dual} and \textit{Mixed} are $0.583$ and $0.600$ respectively. 
\nero{Concerning the \textit{F-score} policy of the \textit{Primal} algorithm, since it does not have corresponding policies for \textit{Mixed} and \textit{Dual} algorithms, we decided to omit all \textit{F-score} plots. 
However, we report the results related to this policy in Table~\ref{tab:results}. 
In the Chemical dataset, this policy does not improve the predictive performance of the \textit{Primal} algorithm, where the \textit{weight} policy is consistently the best performing one.}
Analyzing the plots we can see that the \textit{Primal} algorithm (Algorithm~\ref{alg:explicitbudget}) is not only competitive but it always outperforms \textit{Dual} and \textit{Mixed} in both the \textit{weight} and \textit{oldest} policies. 
Table~\ref{tab:results} shows that, practically in all cases, a higher budget increases the classification performance on the {\textit Chemical} dataset, implying that \textit{Dual} and \textit{Mixed} would probably need a significantly higher budget to reach the performances of \textit{Primal} with $B=10k$. 

Unfortunately, setting $B>50k$ for these algorithms \nero{on the \textit{Chemical} dataset} is unfeasible because of computational times, as it is possible to see from Figure~\ref{fig:TimesDDKChem} reports the average time in seconds needed for the three considered algorithms, instantiated with the $ODD_{ST}$ kernel, to process the \textit{Chemical} dataset with $B=10k$ and $50k$. 

The figure shows that there is a clear gap between the computational times of Algorithms \textit{Primal}, \textit{Mixed} and \textit{Dual}.
Similar considerations can be drawn for NSPDK and FS kernels.  
With budget $10k$, the time needed by the \textit{Primal} algorithm to process a single example is on average ($h=\{0\ldots4\}$) $0.004$ seconds, while for the \textit{Dual} algorithm the required time is $0.2$ seconds. 
The gap grows when setting the budget to $50k$. In this case  the \textit{Primal} algorithm needs on average $0.006$ seconds, while for the \textit{Dual} algorithm already with $h=0$ the required time per example is $0.05$ seconds (almost ten times slower than \textit{Primal}), with h=1 it is $0.39$ seconds. 
With $h=3$ and $4$ the experiments did not complete in $48$ hours, meaning that the processing of each example required more than $1$ second on average. 
The \textit{Mixed} algorithm has computational times similar to the \textit{Primal} ones, but with \rosso{considerably worse} predictive performance.
\begin{figure}
\centering
 \includegraphics[width=0.8\linewidth]{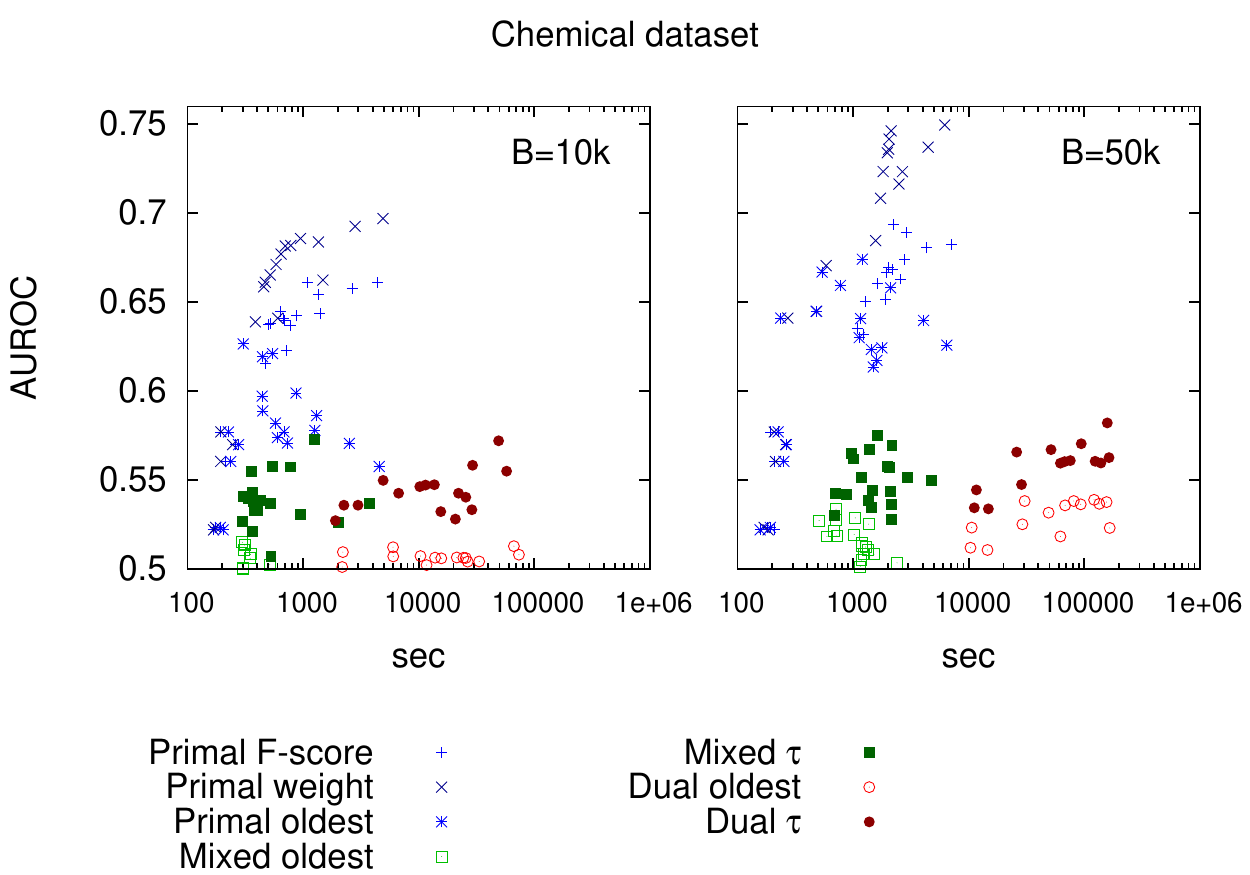}
 \caption{\label{fig:BERTimesChem} Comparison among computational times and AUROC of algorithms \textit{Primal}, \textit{Mixed} and \textit{Dual} on the Image dataset with budget $10k$ and $50k$ for all the considered policies and kernels.}
\end{figure}
%
%

To summarize the results, Figure~\ref{fig:BERTimesChem} shows, for each algorithm, the classification performance in relation to the running time, for budget $10k$ and $50k$. The plots report one point for each algorithm, kernel and parameters combination.
We can see that the \textit{Primal} algorithm has many points in the upper/left part of the plot, meaning that it is able to achieve high predictive performances in a relatively small amount of computational time.
\textit{Mixed} and \textit{Dual} algorithms are all over the lower part of the plot, meaning that they have worse predictive performances and higher running times than \textit{Primal}. 


%
%
\begin{figure}
\centering
 \includegraphics[width=0.8\linewidth]{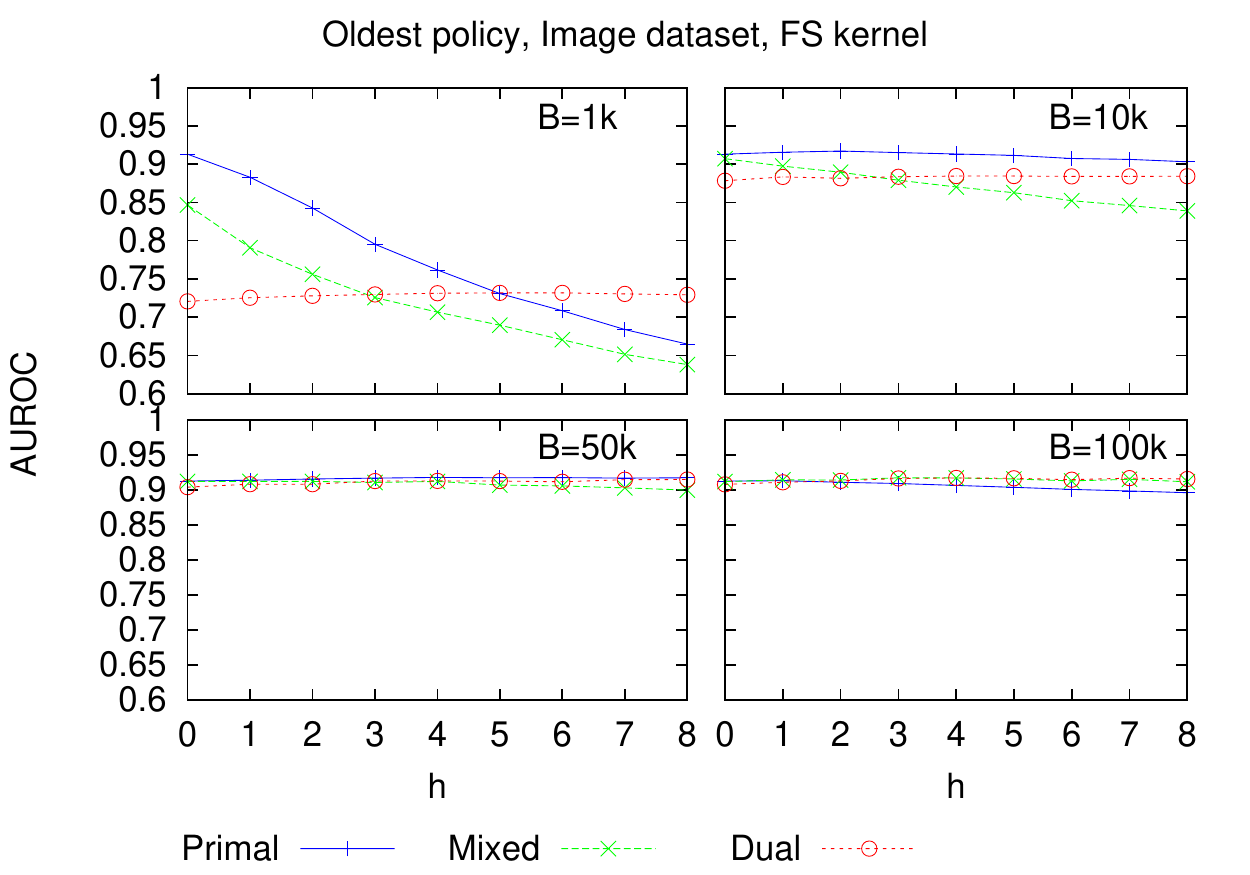}
      \includegraphics[width=0.8\linewidth]{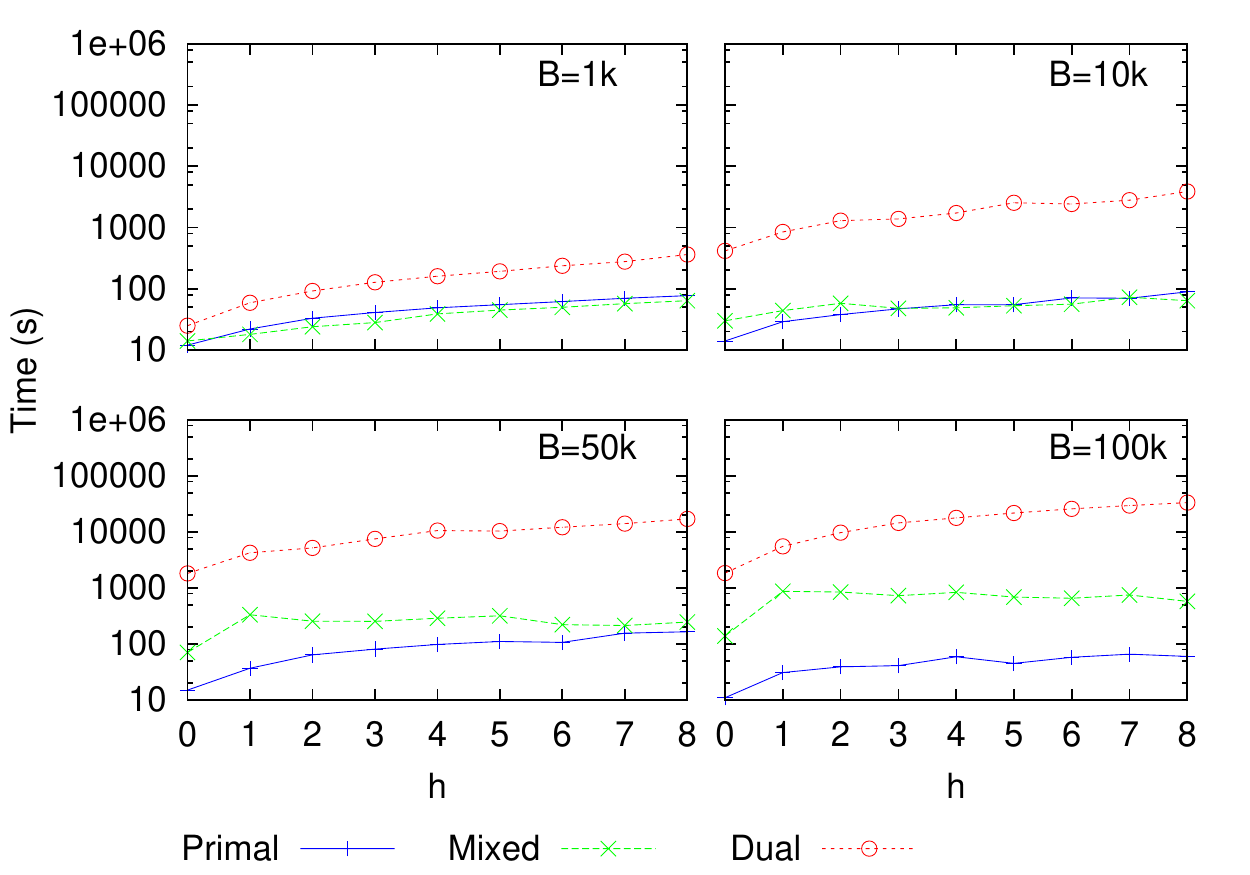}
 \caption{\label{fig:FSKImageoldest} \rosso{Average AUROC value computed over all stream instances for memory budgets $B=1k$ (top left), $B=10k$ (top right), $B=50k$ (bottom left) and $B=100k$ (bottom right) for  algorithms \textit{Primal}, \textit{Mixed} and \textit{Dual} with respect to the values of the $FS$ kernel parameter. Below the first set of plots there is a second one with the corresponding running times}. Plots refer to the \textit{Image} dataset and the \textit{oldest} budget maintainance policy.}
\end{figure}
\begin{figure}
\centering
 \includegraphics[width=0.8\linewidth]{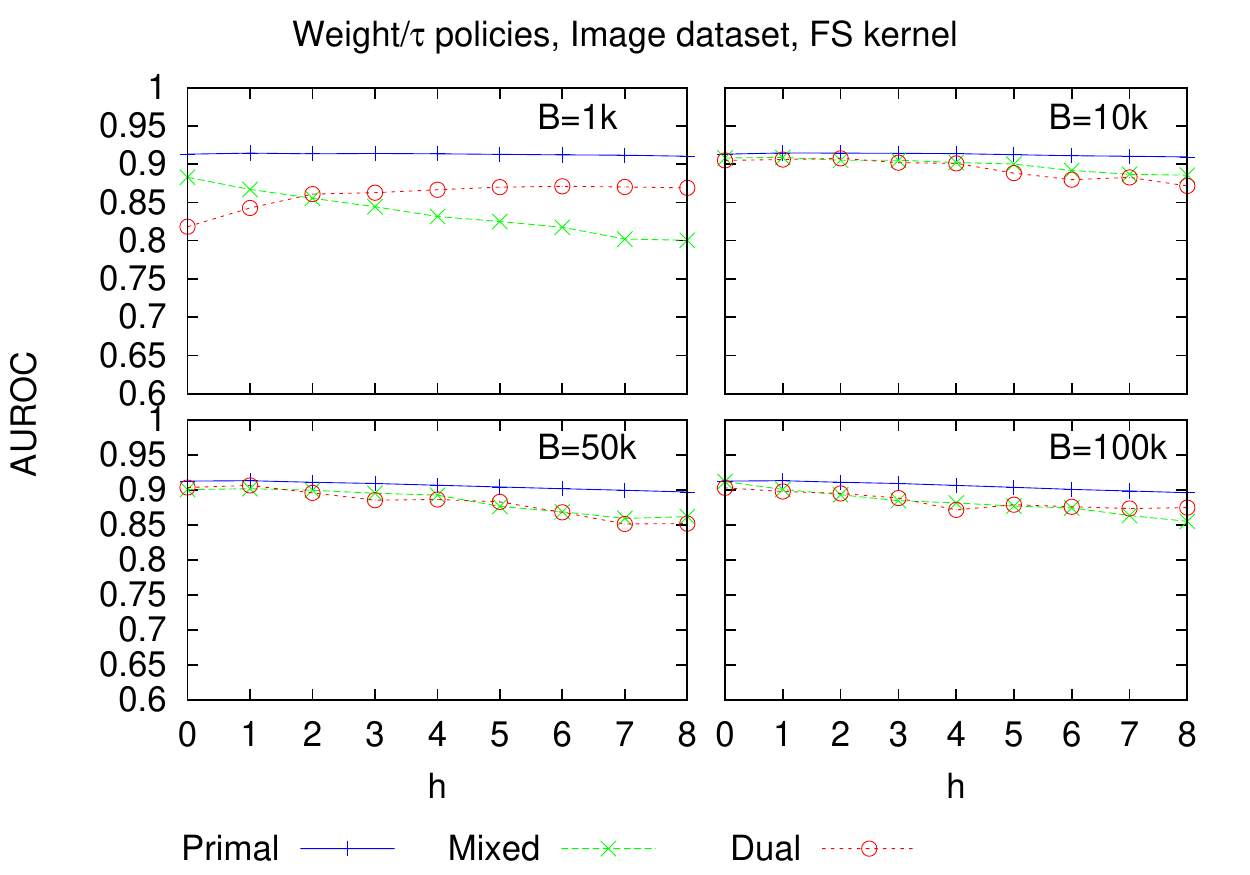}
       \includegraphics[width=0.8\linewidth]{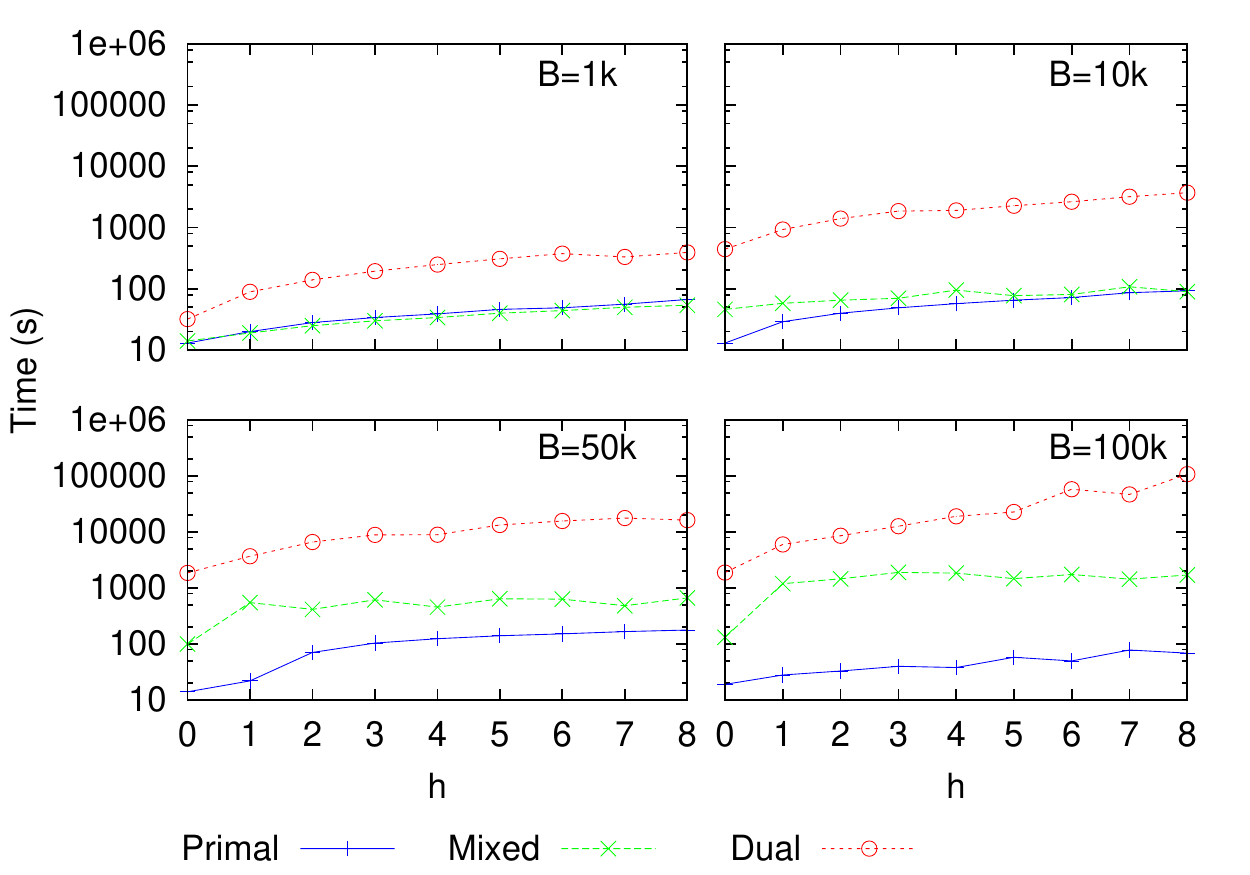}
 \caption{\label{fig:FSKImageweight} \rosso{Average AUROC value computed over all stream instances for memory budgets $B=1k$ (top left), $B=10k$ (top right), $B=50k$ (bottom left) and $B=100k$ (bottom right) for  algorithms \textit{Primal}, \textit{Mixed} and \textit{Dual} with respect to the values of the $FS$ kernel parameter. Below the first set of plots there is a second one with the corresponding running times}. The plots refer to the \textit{Image} dataset and the \textit{weight}/$\tau$ budget maintainance policies.}
\end{figure}
\begin{figure}
\centering
 \includegraphics[width=0.8\linewidth]{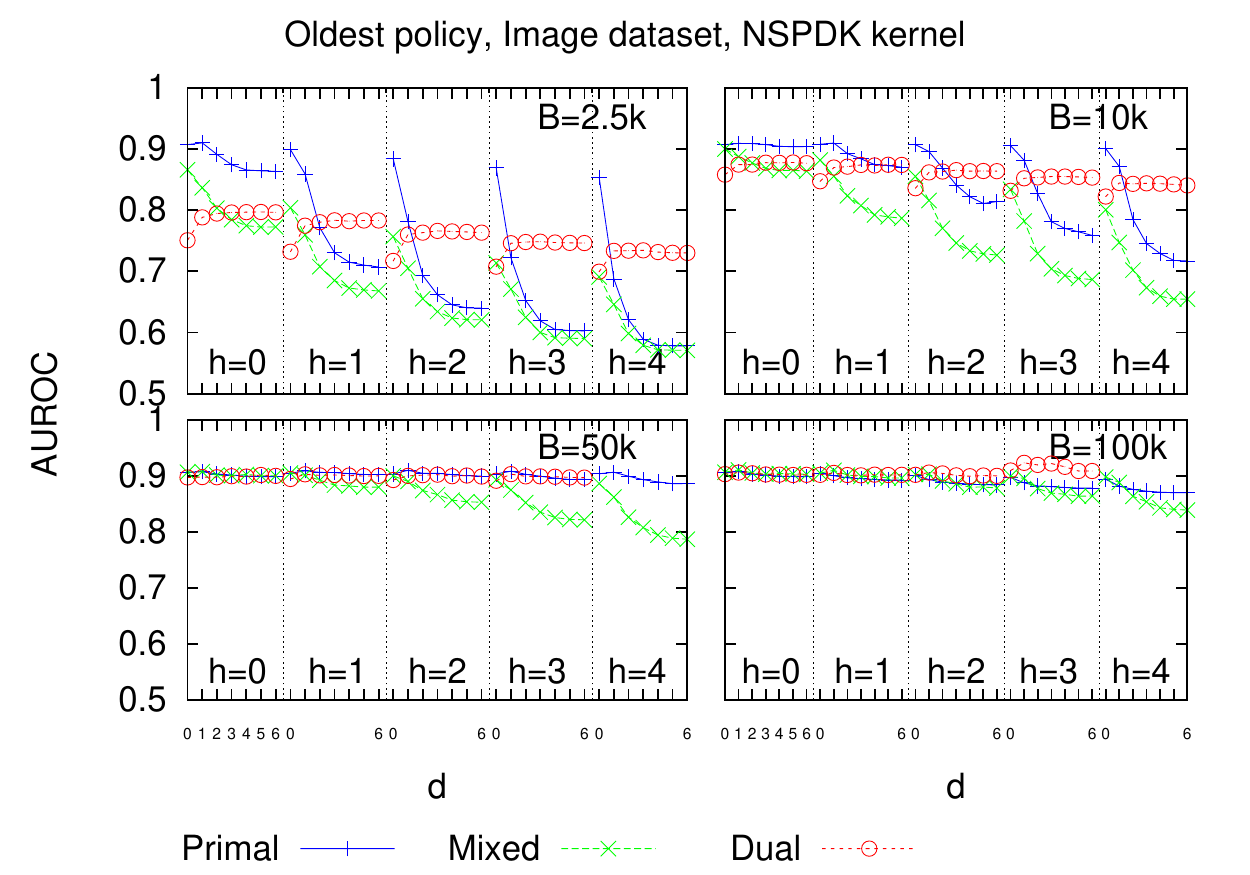}
        \includegraphics[width=0.8\linewidth]{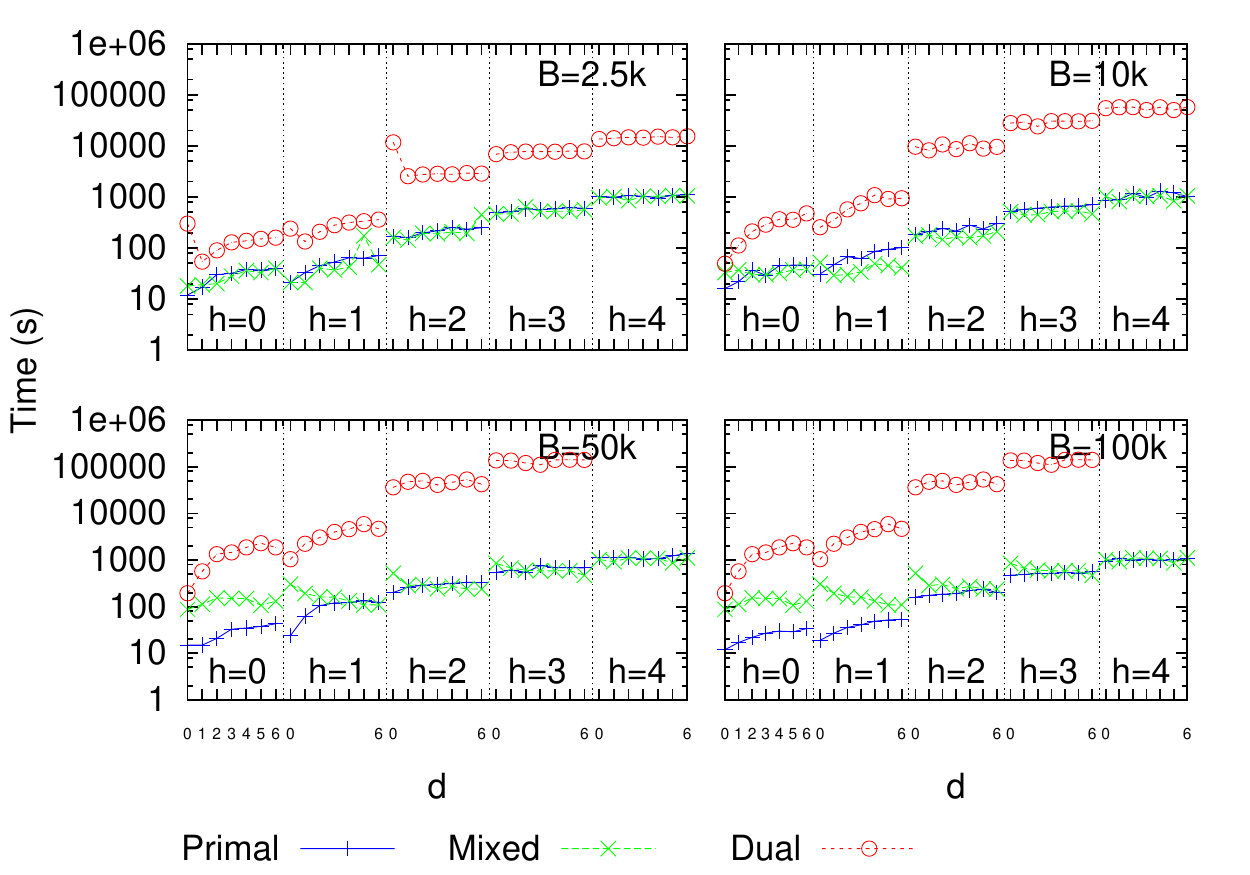}
 \caption{\label{fig:NSPDKImageoldest} \rosso{Average AUROC value computed over all stream instances for memory budgets $B=2.5k$ (top left), $B=10k$ (top right), $B=50k$ (bottom left) and $B=100k$ (bottom right) for  algorithms \textit{Primal}, \textit{Mixed} and \textit{Dual} with respect to the values of the $NSPDK$ kernel parameters. Below the first set of plots there is a second one with the corresponding running times}. The plots refer to the \textit{Image} dataset and the \textit{oldest} budget maintainance policy. \rosso{Missing values indicate that the corresponding execution has not terminated in 48 hours.}}
\end{figure}
\begin{figure}
\centering
 \includegraphics[width=0.8\linewidth]{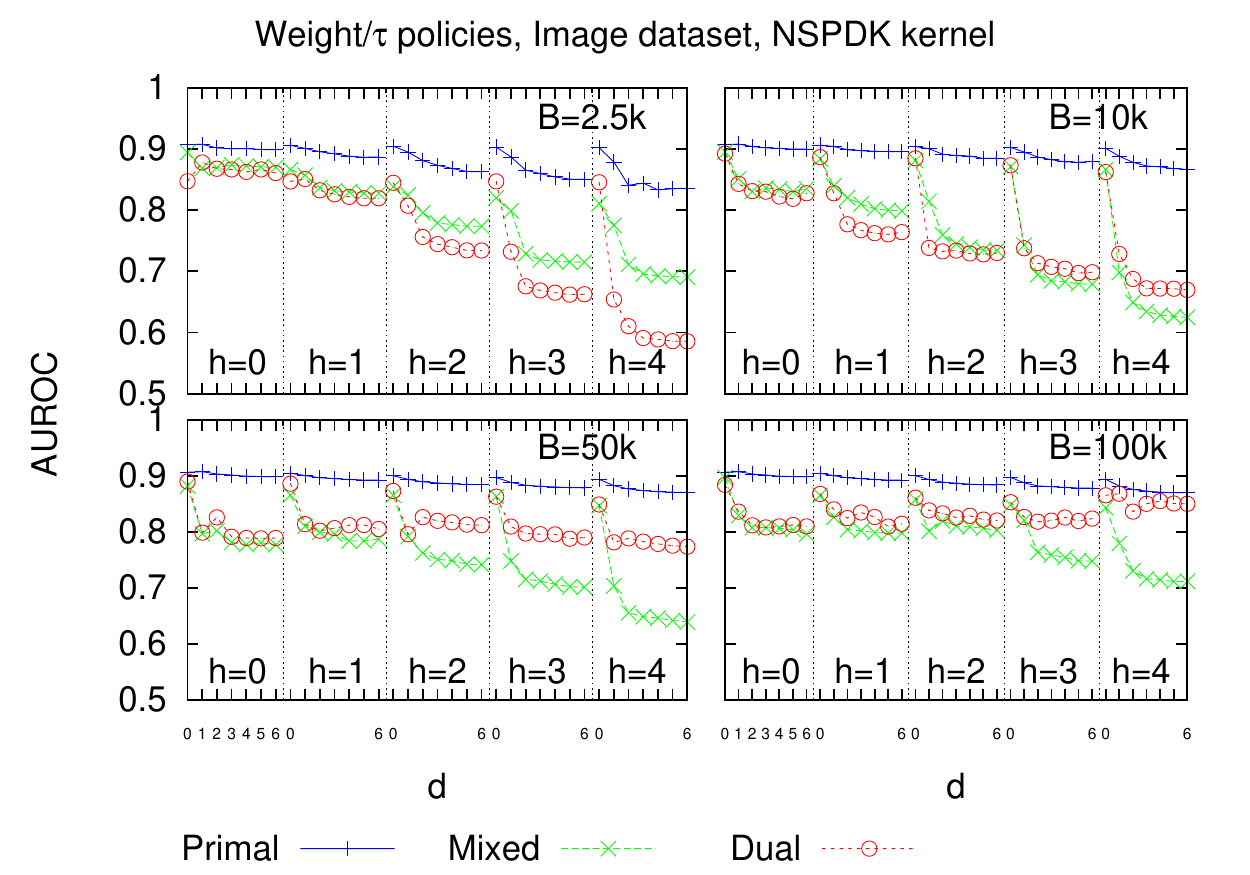}
         \includegraphics[width=0.8\linewidth]{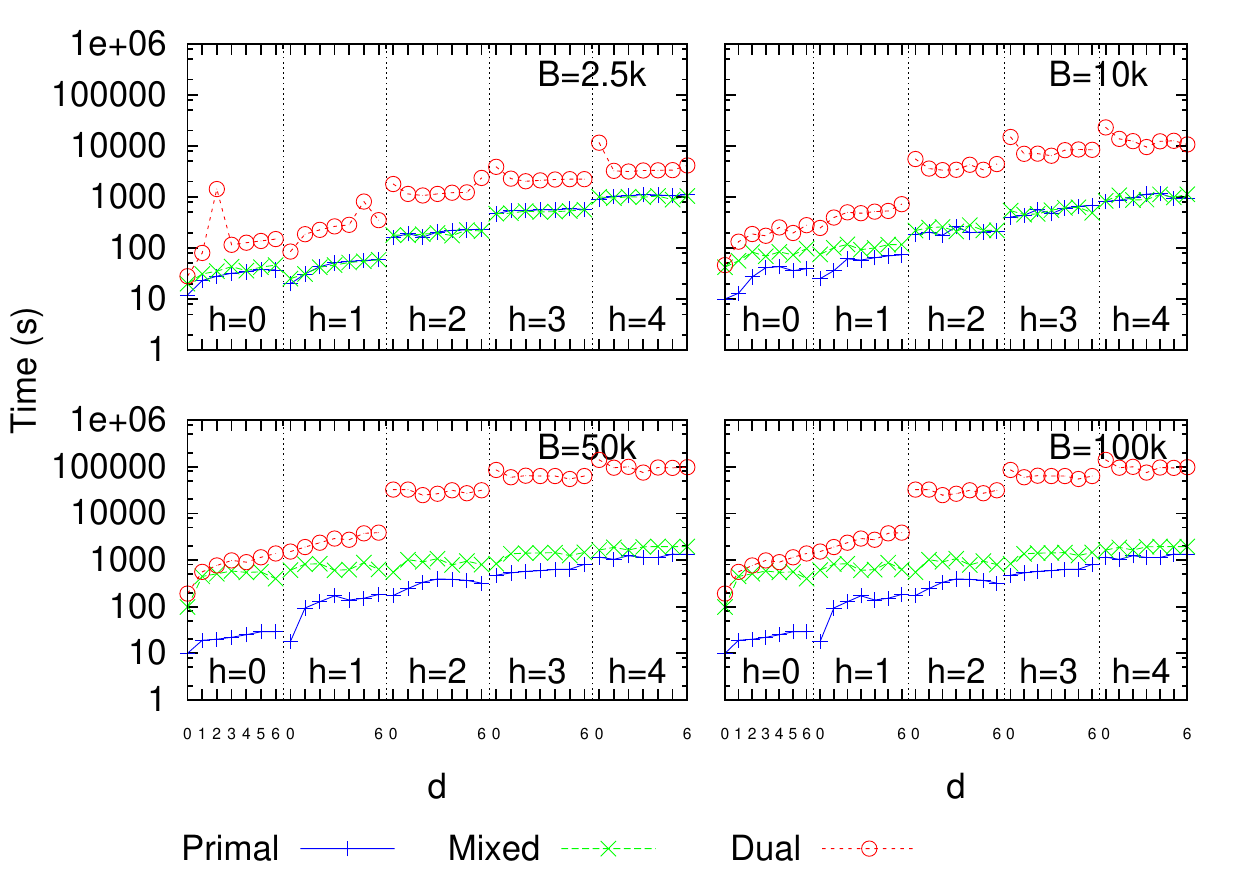}
 \caption{\label{fig:NSPDKImageweight} \rosso{Average AUROC value computed over all stream instances for memory budgets $B=2.5k$ (top left), $B=10k$ (top right), $B=50k$ (bottom left) and $B=100k$ (bottom right) for  algorithms \textit{Primal}, \textit{Mixed} and \textit{Dual} with respect to the values of the $NSPDK$ kernel parameters. Below the first set of plots there is a second one with the corresponding running times}. The plots refer to the \textit{Image} dataset and the \textit{weight}/$\tau$ budget maintainance policies.}
\end{figure}
\begin{figure}
\centering
 \includegraphics[width=0.8\linewidth]{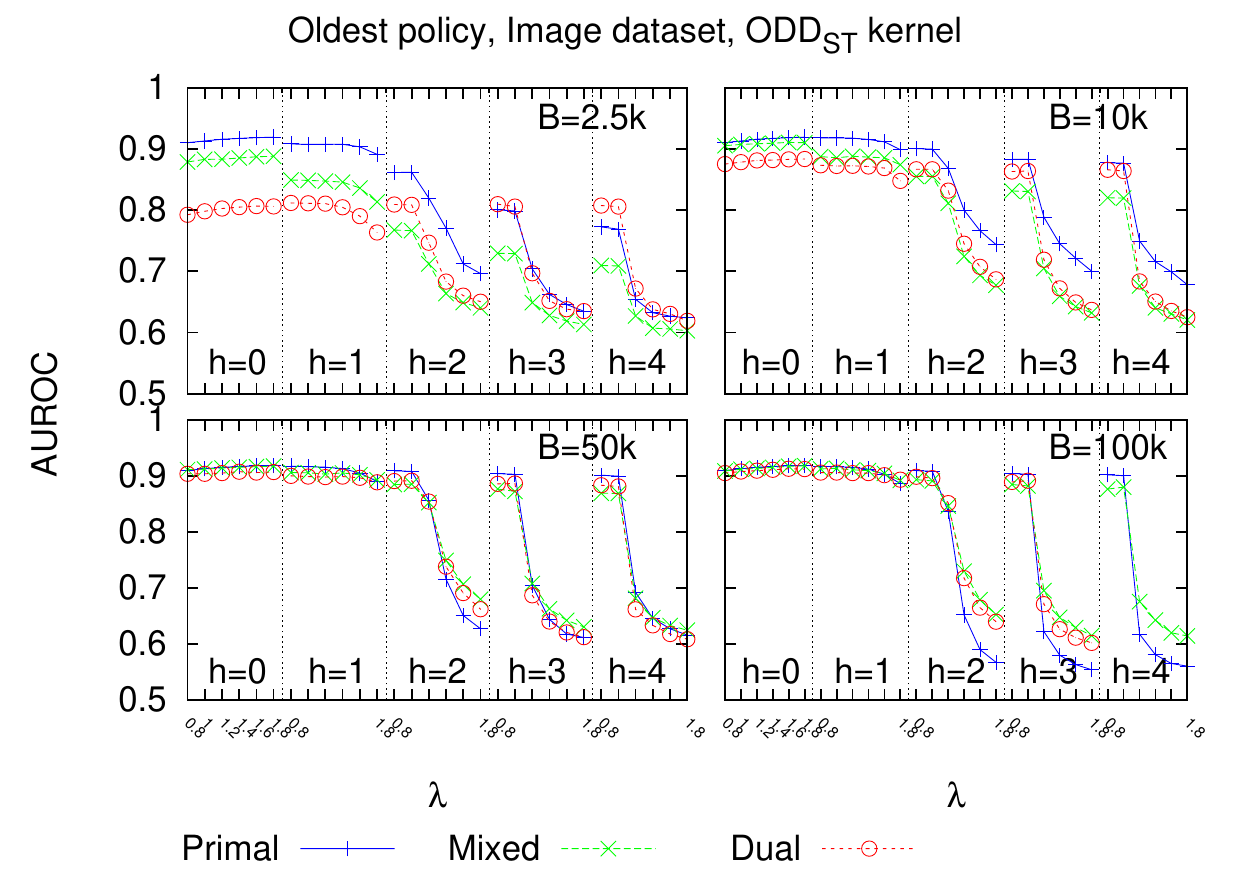}
         \includegraphics[width=0.8\linewidth]{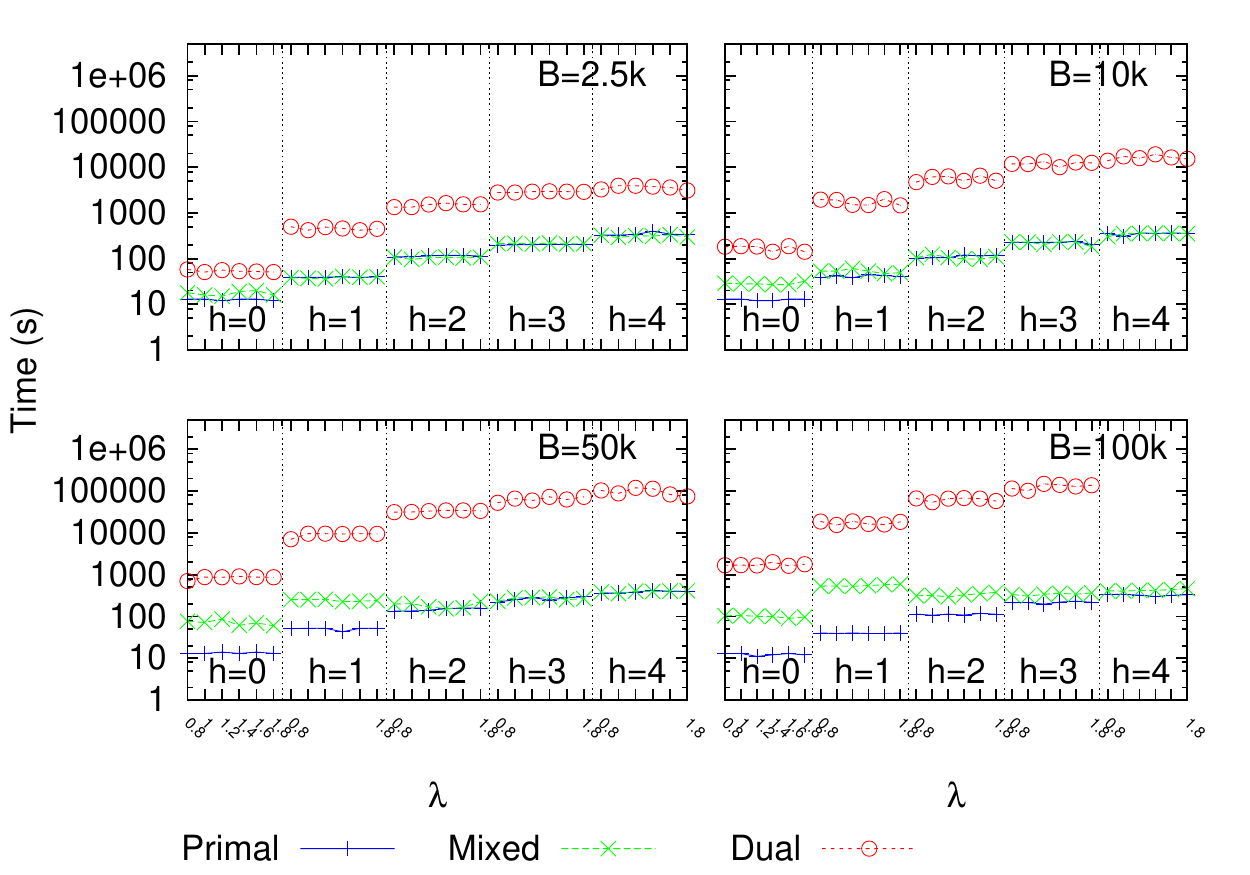}
 \caption{\label{fig:DDKImageoldest} \rosso{Average AUROC value computed over all stream instances for memory budgets $B=2.5k$ (top left), $B=10k$ (top right), $B=50k$ (bottom left) and $B=100k$ (bottom right) for  algorithms \textit{Primal}, \textit{Mixed} and \textit{Dual} with respect to the values of the the $ODD_{ST}$ kernel parameters. Below the first set of plots there is a second one with the corresponding running times}. The plots refer to the \textit{Image} dataset and the \textit{oldest} budget maintainance policy. \rosso{Missing values indicate that the corresponding execution has not terminated in 48 hours.}}
\end{figure}
\begin{figure}
\centering
 \includegraphics[width=0.8\linewidth]{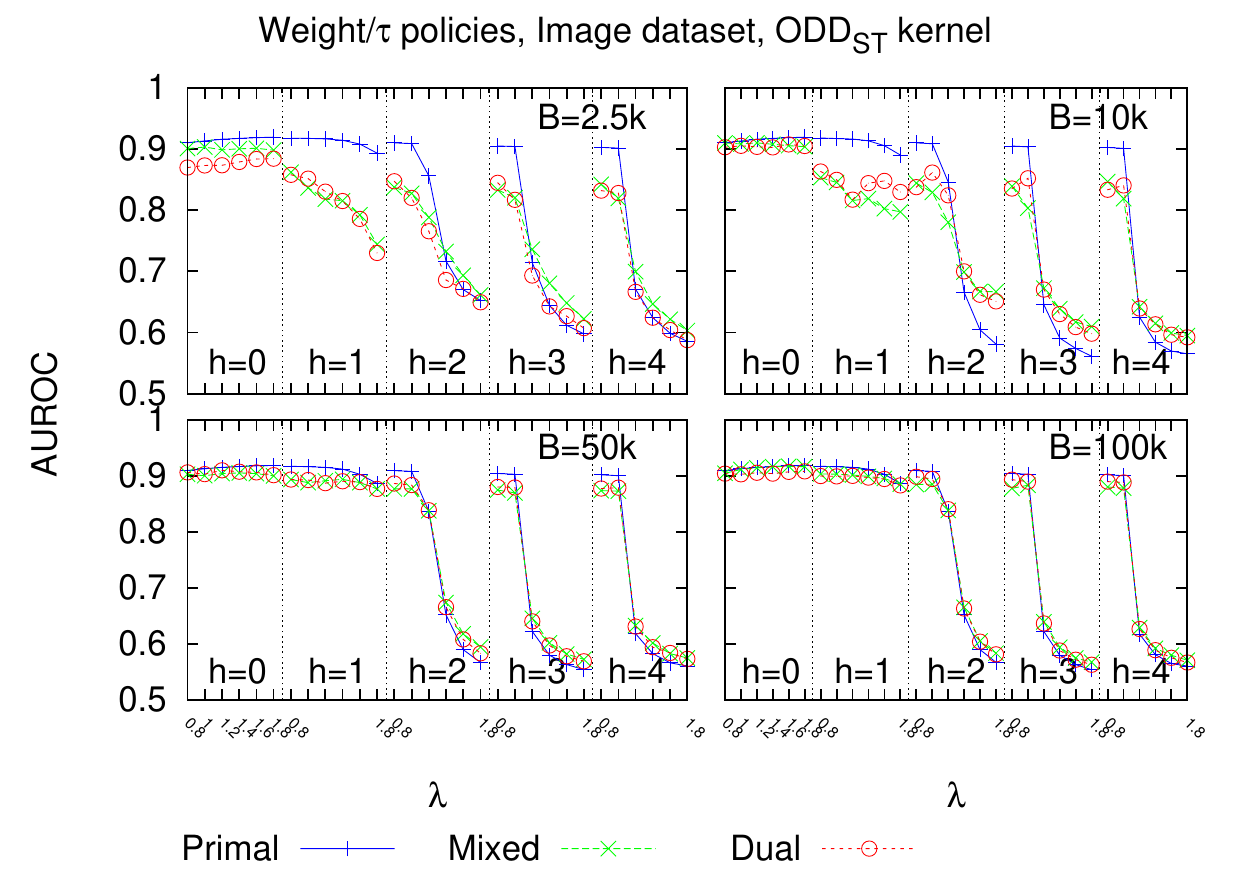}
          \includegraphics[width=0.8\linewidth]{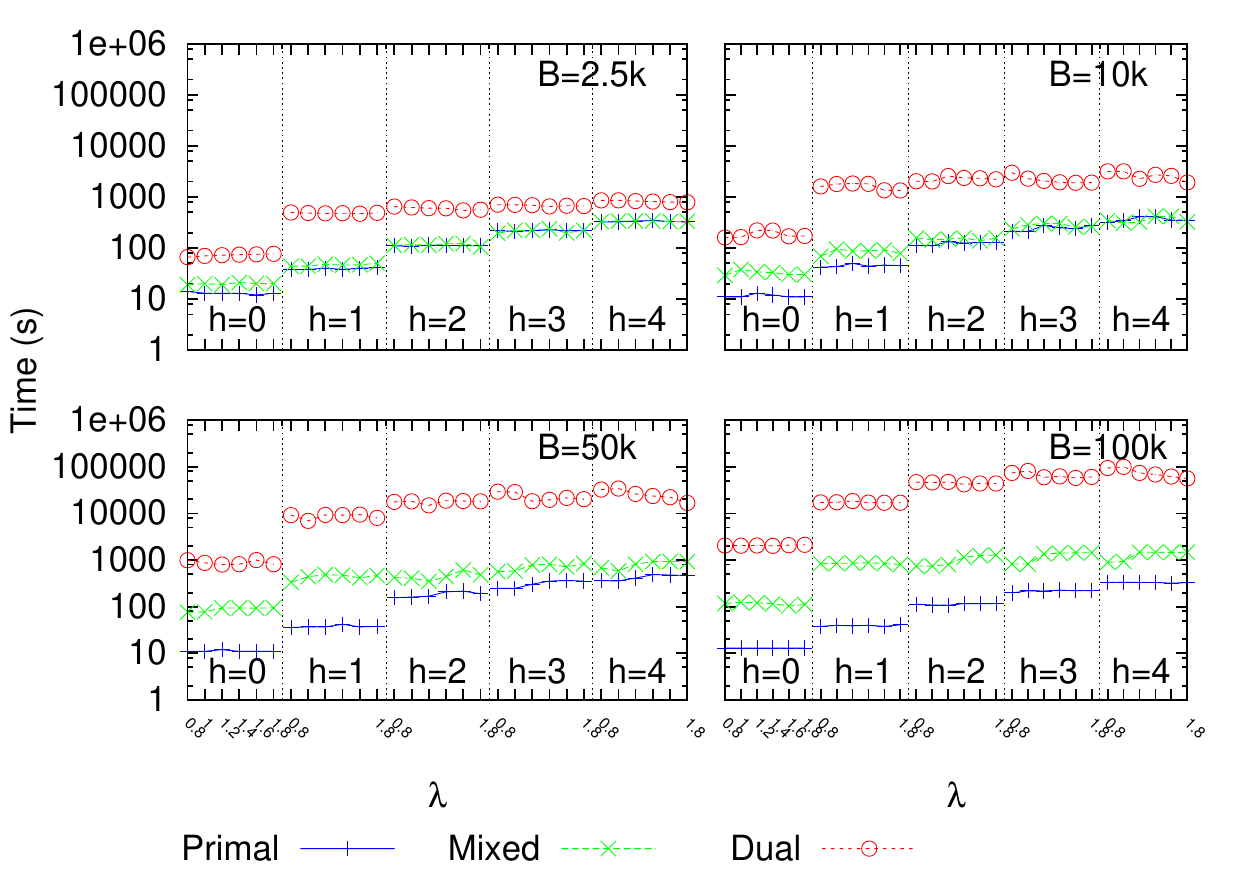}
 \caption{\label{fig:DDKImageweight}\rosso{Average AUROC value computed over all stream instances for memory budgets $B=2.5k$ (top left), $B=10k$ (top right), $B=50k$ (bottom left) and $B=100k$ (bottom right) for  algorithms \textit{Primal}, \textit{Mixed} and \textit{Dual} with respect to the values of the the $ODD_{ST}$ kernel parameters. Below the first set of plots there is a second one with the corresponding running times}. The plots refers to the \textit{Image} dataset and the \textit{weight}/$\tau$ budget maintainance policies.}
 \end{figure}
\begin{figure}
\centering
 \includegraphics[width=0.8\linewidth]{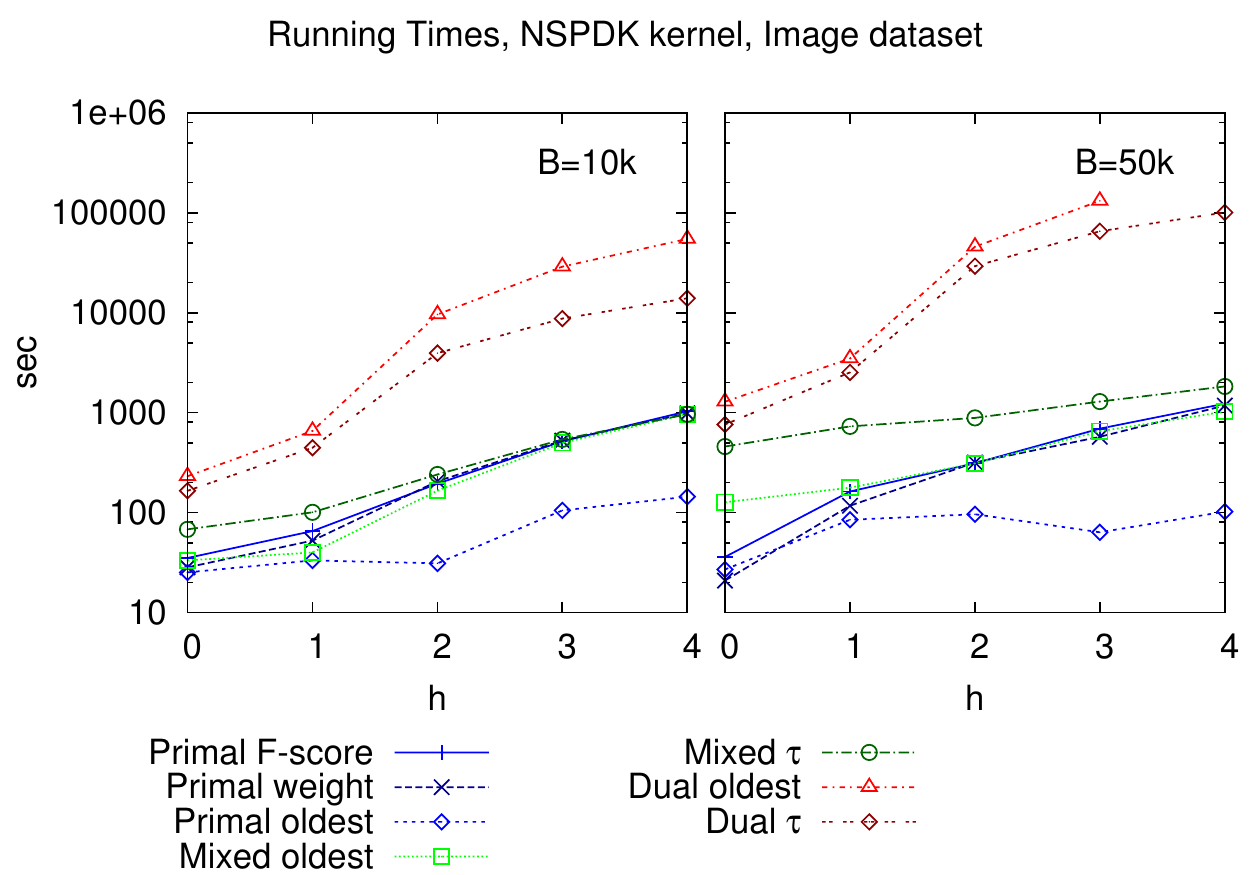}
 \caption{\label{fig:TimesNSPDKiMG} Average computational times of algorithms \textit{Primal}, \textit{Mixed} and \textit{Dual} on the \textit{Image} dataset for the NSPDK kernel.}
\end{figure}

\subsubsection{Experiments on the Image Datasets} \label{sec:expsimage}

The same experimental setting described for the \textit{Chemical} dataset is replicated here for the \textit{Image} dataset. 
%
 Figures~\ref{fig:FSKImageoldest}-\ref{fig:DDKImageweight} show\nero{, for each set of corresponding management policies,} the performance of the kernels with respect to their parameters. 
\nero{We tested different values for the budget size, ranging from $1k$ to $100k$.
In Figure~\ref{fig:FSKImageoldest} we can see that, for small budget values, the \textit{Primal} algorithm is the best performing one with the \textit{oldest} budget management policy. When the budget grows (i.e. for $B=100k$) \textit{Mixed} and \textit{Dual} perform slightly better than \textit{Primal}.
Figure~\ref{fig:FSKImageweight}, referring to the same kernel with \textit{weight} policy, depicts a similar scenario. In this case, \rosso{\textit{Primal} performs slightly better than \textit{Dual} and \textit{Mixed} in all the considered budget sizes.} }

\nero{In Figures~\ref{fig:NSPDKImageoldest} and \ref{fig:NSPDKImageweight} we started from a budget value of $B=2.5k$, since the NSPDK generates more features than FS (as detailed in Section~\ref{sec:oddkernel}).
 When considering the \textit{oldest} policy, \textit{Primal} performs best for budget values up to $10k$. 
 In the case of \textit{weight} policy, \rosso{\textit{Primal} always performs better than \textit{Dual} and \textit{Mixed}.}
  More in general, it is possible to see that the performance of \textit{Dual} and \textit{Mixed} increase proportionally to the budget, while \textit{Primal} performs best with budget $10k$, thus its performance do not improve if more budget is available (note nonetheless that the performance do not decrease significantly).
Apparently, in the case of FS and NSPDK kernels, the classification performances of the different algorithms depend critically on the budget size.
 }
\nero{Figure~\ref{fig:DDKImageoldest}  analyzes the situation with $ODD_{ST}$ kernel and \textit{oldest} policy.
\rosso{Also in this case, \textit{Primal} algorithm is the better performing one with every budget value.
 However, with higher budgets, the other algorithms show comparable performances. Also in this case, the higher the budget the better the predictive performances of \textit{Mixed} and \textit{Dual}.}
 The scenario is similar when cosidering the \textit{weight}/$\tau$ policies in Figure \ref{fig:DDKImageweight}. 
}
The running times of the different kernels on the \textit{Image} dataset are in general lower with respect to the \textit{Chemical} one. 
Figure~\ref{fig:TimesNSPDKiMG} reports the running time required by the FS kernel with budget $10,000$. As for the \textit{Chemical} dataset the \textit{Primal} and \textit{Mixed} algorithms are way faster that the \textit{Dual} algorithm. 
%
\begin{figure}
\centering
 \includegraphics[width=0.8\linewidth]{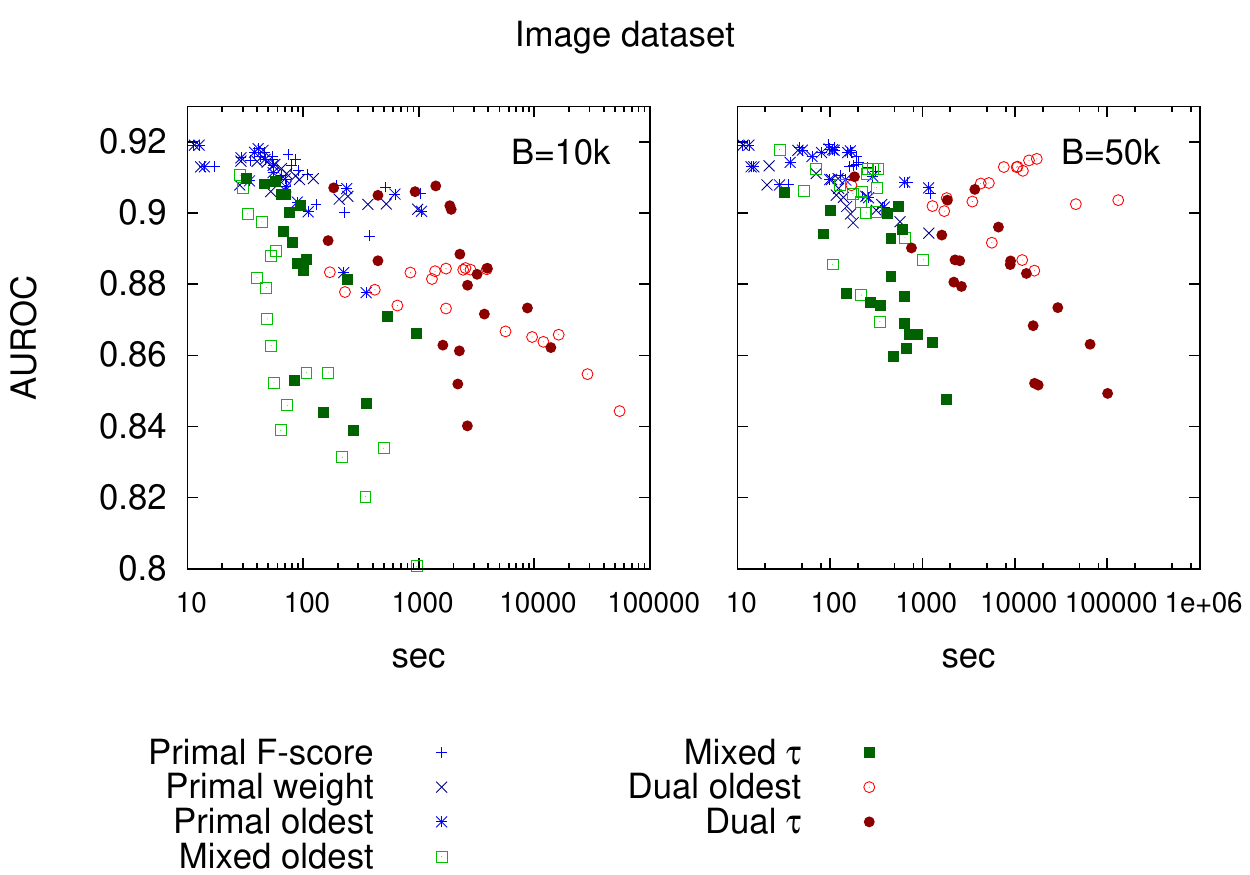}
 \caption{\label{fig:BERTimesImage} Comparison among computational times and AUROC of algorithms \textit{Primal}, \textit{Mixed} and \textit{Dual}on the Image dataset with budget $10k$ and $50k$ for all the considered policies and kernels.}
\end{figure}
%

Figure~\ref{fig:BERTimesImage} shows the predictive performance in relation to the computational time required from the different algorithms in the \textit{Image} dataset. 
The \textit{Primal} algorithm is the fastest, with some points at the leftmost margin of the plots. Also from a predictive performance point of view, we see that the algorithm with the highest AUROC is \textit{Primal} for both budget values .With $B=50k$ the \textit{Mixed} and \textit{Dual} algorithms achieve similar performances, although with a higher runtime.

To summarize, given a budget management policy, under a certain budget size \textit{Primal} algorithm is the best performing one, and over that size \textit{Dual} and \textit{Primal} (and in some cases \textit{Mixed}) show very similar performances. However, there is a significant difference in the computational times required by the different algorithms, with \textit{Primal} and \textit{Mixed} being considerably faster than \textit{Dual}. 


\subsubsection{Discussion} \label{sec:expsdiscussion}

We can draw some final remarks concluding our experimental analysis. 
First it is worth to point out that our analysis refers only to those kernels which allow for an explicit feature space representation. 
Such kernels are only a subset of the existing graph kernels. However, they are the ones currently having state-of-the-art predictive performances. 
%
\nuovo{While the \textit{Dual} algorithm can represent more compactely the model than the \textit{Primal} approach when the feature space associated to the kernel is very large, this implies a loss in efficiency when computing the score for a new graph: the kernel value between the input graph  and all the graphs in the model have to be computed from scratch. 
As the values of Figures~\ref{fig:TimesDDKChem} and~\ref{fig:TimesNSPDKiMG} indicate, that makes the application of the \textit{Dual} algorithm to graph streams practically infeasible, especially when strict time constraints have to be satisfied. 
 The \textit{Mixed} algorithm is able to significantly speed up the score computation by storing the explicit feature space representation of each graph in the model. As a consequence, the size of the model may increase significantly, thus reducing the total number of graphs that can be kept in it: \textit{Dual} algorithm is able to store in memory approximately $250$ graphs of the chemical datasets with budget $10,000$, while \textit{Mixed} algorithm only $100$ graphs.  
On the contrary, \textit{Primal} algorithm keeps in the model only the most informative features, and thus it is able to retain information of all graphs inserted in the model while preserving a very good efficiency. 
According to our experiments, there is a budget value which determines whether the \textit{Primal} or the other approaches are preferable. 
While such threshold value can be observed in our experiments for the \textit{Image} dataset,
due to the inefficiency of \textit{Dual} and \textit{Mixed}, we were not able to identify it for the \textit{Chemical} dataset (where \textit{Primal} always outperforms the other approaches). }

\section{Conclusions and Future Work}\label{sec:conclusions}
In this work we analyzed the trade-off between efficiency and efficacy of various
versions of online margin kernel perceptron algorithms  when dealing with graph streams and under the assumption of fixed memory budgets. 
One of them
efficiently exploits the explicit representation of the feature space (via hash tables) of different state-of-the-art graph kernels recently defined in literature. 

Experimental results on real-world datasets show that, under a threshold budget size, working in feature space is preferable both in terms of classification performance and running times. In a future work we will investigate the dependency between such budget value and the size of the feature space
associated to the kernel, the policy for pruning the model and the nature of the dataset. 

\section{Acknowledgments}
This work was supported by the University of Padova under the strategic project \textit{BIOINFOGEN}.

\section*{References}
\bibliographystyle{elsarticle-num-names} 
\bibliography{other,swork,ijcai11,biblography}

\end{document}